\documentclass{article}

\usepackage{microtype}
\usepackage{graphicx}
\usepackage{subfigure}
\usepackage{booktabs} 
\usepackage{hyperref}

\usepackage{amsmath,amsfonts,amssymb,amsthm,mathabx}
\usepackage{dsfont}
\usepackage{comment}
\usepackage[export]{adjustbox}
\usepackage{xcolor}
\usepackage{rotating}
\usepackage{todonotes}

\usepackage[accepted]{icml2019}

\newcommand{\classifier}{F}
\newcommand{\logit}{f}
\newcommand{\noise}{{\mathbb N}}
\newcommand{\datad}{{\mathbb P}}

\usepackage{amsmath,amsfonts,bm}

\def\eqref#1{Equation~\ref{#1}}
\def\Eqref#1{Equation~\ref{#1}}
\def\1{\bm{1}}

\newcommand{\E}{{\mathbf E}}

\newtheorem{proposition}{Proposition}

\DeclareMathAlphabet{\mathsfit}{\encodingdefault}{\sfdefault}{m}{sl}
\SetMathAlphabet{\mathsfit}{bold}{\encodingdefault}{\sfdefault}{bx}{n}

\graphicspath{{./figures/}}

\icmltitlerunning{The Odds are Odd: A Statistical Test for Detecting Adversarial Examples}

\begin{document}

\twocolumn[
\icmltitle{The Odds are Odd: \\A Statistical Test for Detecting Adversarial Examples}

\icmlsetsymbol{equal}{*}

\begin{icmlauthorlist}
\icmlauthor{Kevin Roth}{equal,eth}
\icmlauthor{Yannic Kilcher}{equal,eth}
\icmlauthor{Thomas Hofmann}{eth}
\end{icmlauthorlist}

\icmlaffiliation{eth}{Department of Computer Science, ETH Z\"urich}

\icmlcorrespondingauthor{}{kevin.roth@inf.ethz.ch}
\icmlcorrespondingauthor{}{yannic.kilcher@inf.ethz.ch}
\icmlcorrespondingauthor{}{thomas.hofmann@inf.ethz.ch}

\icmlkeywords{Detecting Adversarial Perturbations, Adversarial Robustness, Deep Learning}

\vskip 0.3in
]

\printAffiliationsAndNotice{\icmlEqualContribution}

\begin{abstract}
We investigate conditions under which test statistics exist that can reliably detect examples, which have been adversarially manipulated in a white-box attack. These statistics can be easily computed and calibrated by randomly corrupting inputs. 
They exploit certain anomalies that adversarial attacks introduce, in particular if they follow the paradigm of choosing perturbations optimally under $p$-norm constraints. Access to the log-odds is the only requirement to defend models.  We justify our approach empirically, but also provide conditions under which detectability via the suggested test statistics is guaranteed to be effective. In our experiments, we show that it is even possible to correct test time predictions for adversarial attacks with high accuracy.
\end{abstract}

% !TEX root = 00_icml2019.tex

\section{Introduction}
\label{sec:introduction}

Deep neural networks have been used with great success for perceptual tasks such as image classification \cite{simonyan2014very, lecun2015deep} or speech recognition \cite{hinton2012deep}. 
While they are known to be robust to random noise, it has been shown that the accuracy of deep nets can dramatically deteriorate in the face of  so-called adversarial examples \cite{biggio2013evasion, szegedy2013intriguing, goodfellow2014explaining}, i.e.\ small perturbations of the input signal, often imperceptible to humans, that are sufficient to induce large changes in the model output. 

A plethora of methods have been proposed to find adversarial examples \cite{szegedy2013intriguing, goodfellow2014explaining, kurakin2016adversarial, moosavi2016deepfool, sabour2015adversarial}. 
These often transfer across different architectures, enabling black-box attacks even for inaccessible models \cite{papernot2016transferability, kilcher2017best, tramer2017space}.
This apparent vulnerability is worrisome as deep nets  start to proliferate in the real-world, including in safety-critical deployments.

The most direct and popular strategy of robustification is to use adversarial examples  as data augmentation during training \cite{goodfellow2014explaining, kurakin2016adversarial, madry2017towards},
which improves robustness against specific attacks, yet does not address vulnerability to more cleverly designed counter-attacks \cite{athalye2018obfuscated, carlini2017adversarial}. This raises the question of whether one can protect models with regard to a wider range of possible adversarial perturbations.

A different strategy of defense is to detect whether or not the input has been perturbed,
by detecting characteristic regularities either in the adversarial perturbations themselves or in the network activations they induce \cite{grosse2017statistical, feinman2017detecting, xu2017feature, metzen2017detecting, carlini2017adversarial}. In this spirit, we propose a method  that measures how feature representations and log-odds change under noise: If the input is adversarially perturbed, the noise-induced feature variation tends to have a characteristic direction, whereas it tends not to have any specific direction if the input is natural.  We evaluate our method against strong iterative attacks and show that even an adversary aware of the defense cannot evade our detector.
E.g.\ for an $L^\infty$-PGD white-box attack on CIFAR10, our method achieves a detection rate of $99\%$ (FPR $<1\%$), with accuracies of $96\%$ on clean and $92\%$ on adversarial samples respectively. 
On ImageNet, we achieve a detection rate of $99\%$ (FPR $1\%$).
Our code can be found at \url{https://github.com/yk/icml19_public}.

In summary, we make the following contributions: 
\vspace{-2mm}
\begin{itemize}
\item We propose a statistical test for the detection and classification of adversarial examples.
\item We establish a link between adversarial perturbations and inverse problems, providing valuable insights into the feature space kinematics of adversarial attacks.
\item We conduct extensive performance evaluations as well as a range of experiments to shed light on aspects of adversarial perturbations that make them detectable.  
\end{itemize}

% !TEX root = 00_icml2019.tex

\section{Related Work}
\label{sec:background}

{\bf Iterative adversarial attacks.}
Adversarial perturbations are small specifically crafted perturbations of the input, typically imperceptible to humans, that are sufficient to induce large changes in the model output.
Let $f$ be a probabilistic classifier with logits $f_y$ and let $F(x) = \arg\max_{y} f_y(x)$.
The goal of the adversary is to find an $L^p$-norm bounded perturbation $\triangle x \in \mathcal{B}^p_{\epsilon}(0) := \{ \triangle : || \triangle ||_p \leq \epsilon \}$, 
where $\epsilon$ controls the attack strength, 
such that the perturbed sample $x+\triangle x$ gets misclassified by the classifier $F(x)$.
Two of the most iconic iterative adversarial attacks are: 

Projected Gradient Descent \cite{madry2017towards} aka Basic Iterative Method \cite{kurakin2016adversarial}: 
\begin{align}
x^0  &\sim \mathcal{U}( \mathcal{B}^p_{\epsilon}(x) ) \\
x^{t\hspace{-1pt}+\!1} &= \Pi_{\mathcal{B}^\infty_{\epsilon}(x) } \left(  x^t \!-\! \alpha\,{\rm sign}(\left.\nabla_x \mathcal{L}(f; x, y)\right|_{x^t} )  \right) \,\,[L^\infty]\nonumber \\
x^{t\hspace{-1pt}+\!1} &= \Pi_{\mathcal{B}^2_{\epsilon}(x) } \Big(  x^t \!-\! \alpha \frac{ \left.\nabla_x \mathcal{L}(f; x, y)\right|_{x^t} }{|| \left.\nabla_x \mathcal{L}(f; x, y)\right|_{x^t} ||_2}  \Big) \quad\ [L^2]\nonumber 
\end{align}
where the second and third line refer to the $L^\infty$- and $L^2$-norm variants respectively, 
$\Pi_\mathcal{S}$ is the projection operator onto the set $\mathcal{S}$, 
$\alpha$ is a small step-size, $y$ is the target label
and $\mathcal{L}(f; x, y)$ is a suitable loss function. 
For untargeted attacks $y = F(x)$ and the sign in front of $\alpha$ is flipped, so as to ascend the loss function.

Carlini-Wagner attack \cite{carlini2017towards}: 
\vspace{-1pt}
\begin{equation}
\begin{aligned}
{\rm minimize} \,\,\,\,\,  & || \triangle x ||_p + c \mathcal{F}(x + \triangle x) \\
{\rm such\,\,that} \,\,\,\, & x + \triangle x \in {\rm dom}_x \\[-3pt]
\end{aligned}
\end{equation}
where $\mathcal{F}$ is an objective function, defined such that $\mathcal{F}(x + \triangle x) \leq 0$ if and only if $F(x+\triangle x) = y$, e.g.\ 
$\mathcal{F}(x) = \max( \max\{ f_z(x) : z\neq y \} - f_y(x), -\kappa )$
(see Section~V.A in \cite{carlini2017towards} for a list of objective functions with this property) and ${\rm dom}_x$ denotes the data domain, e.g.\ ${\rm dom}_x = [0,1]^D$. The constant $c$ trades off perturbation magnitude (proximity) with perturbation strength (attack success rate) and is chosen via binary search.

{\bf Detection.} 
The approaches most related to our work are those that defend a machine learning model against adversarial attacks by detecting whether or not the input has been perturbed, either by detecting characteristic regularities in the adversarial perturbations themselves or in the network activations they induce \cite{grosse2017statistical, feinman2017detecting, xu2017feature, metzen2017detecting, song2017pixeldefend, li2017adversarial, lu2017safetynet, carlini2017adversarial}.

Notably, \citet{grosse2017statistical} argue that adversarial examples are not drawn from the same distribution as the natural data and can thus be detected using statistical tests.
\citet{metzen2017detecting} propose to augment the deep classifier net with a binary ``detector'' subnetwork that gets input from intermediate feature representations and is trained to discriminate between natural and adversarial network activations.
\citet{feinman2017detecting} suggest to detect adversarial examples by testing whether inputs lie in low-confidence regions of the model either via kernel density estimates in the feature space of the last hidden layer or via dropout uncertainty estimates of the classfier's predictions.
\citet{xu2017feature} propose to detect adversarial examples by comparing the model's predictions on a given input with its predictions on a squeezed version of the input, such that if the difference between the two exceeds a certain threshold, the input is considered to be adversarial.
A quantitative comparison with the last two methods can be found in the Experiments~Section.

{\bf Origin.} 
It is still an open question whether adversarial examples exist because of intrinsic flaws of the model or learning objective or whether they are solely the consequence of non-zero generalization error and high-dimensional statistics \cite{gilmer2018adversarial, schmidt2018adversarially, fawzi2018adversarial}.
We note that our method works regardless of the origin of adversarial examples:
as long as they induce characteristic regularities in the feature representations of a neural net, e.g.\ under noise, they can be detected.

% !TEX root = 00_icml2019.tex

\section{Identifying and Correcting Manipulations}
\label{sec:method}

\subsection{Perturbed Log-Odds}

We work in a multiclass setting, where pairs of inputs $x^* \in \Re^D$ and class labels $y^* \in \{1,\dots,K\}$ are generated from a data distribution $\datad$. The input may be subjected to an adversarial perturbation $x = x^* + \triangle x$ such that $\classifier(x)  \neq y^* = \classifier(x^*)$, forcing a  misclassification.
A well-known defense strategy against such manipulations is to voluntarily corrupt inputs by noise before processing them. The rationale is  that by adding noise $\eta \sim \noise$, one may be able to recover the original class, if $\text{Pr}\left\{ \classifier(x + \eta) = y^* \right\}$ is  sufficiently large. For this to succeed, one typically utilizes domain knowledge in order to construct meaningful families of random transformations, as has been demonstrated, for instance, in \cite{xie2017mitigating, athalye2017synthesizing}. Unstructured (e.g.~white) noise, on the other hand, does typically not yield practically viable tradeoffs between probability of recovery and overall accuracy loss. 

We thus propose to look for more subtle statistics that can be uncovered by using noise as a \textit{probing instrument} and not as a direct means of recovery. We will focus on probabilistic classifiers with a logit layer of scores as this gives us access to continuous values.  For concreteness we will explicitly parameterize logits via $\logit_y(x) = \langle w_y, \phi(x)\rangle$ with class-specific weight vectors $w_y$ on top of a feature map $\phi$ realized by a (trained) deep network. Note that typically $F(x) = \arg\max_{y} f_y(x)$. We also define pairwise log-odds between classes $y$ and $z$, given input $x$
\begin{align}
\logit_{y,z}(x) &= f_z(x) - f_y(x) = \langle w_z \!-\! w_y, \phi(x) \rangle\,.
\end{align}
We are interested in the noise-perturbed log-odds $f_{y,z}(x+\eta)$ with $\eta \sim \noise$, where $y = y^*$, if ground truth is available, e.g.~during training, or $y = F(x)$, during testing. 

Note that the log-odds may behave differently for different class pairs, as they reflect class confusion probabilities  that are task-specific and that cannot be anticipated \textit{a priori}. 
This can be addressed by performing a Z-score standardization across data points $x$ and perturbations $\eta$. For each fixed class pair $(y,z)$ define:
\begin{equation}
\begin{aligned}
g_{y,z}(x, \eta) & := f_{y,z}(x+\eta) - f_{y,z}(x) \\
\mu_{y^*,z} & := \mathbf E_{x^* |y^*} \mathbf E_{\eta} \left[ g_{y^*,z}(x^*, \eta)  \right] \\
\sigma^2_{y^*,z} & := \mathbf E_{x^*|y^*} \mathbf E_{\eta} \left[ (g_{y^*,z}(x^*, \eta)-\mu_{y^*,z}) ^2 \right] \\
\bar g_{y,z}(x, \eta) & := \left[ g_{y,z}(x, \eta) - \mu_{y,z} \right] / \sigma_{y,z}\,.
\end{aligned}
\end{equation}
In practice, all of the above expectations are computed by sample averages over training data and noise instantiations. 
Also, note that $g_{y,z}(x, \eta) = \langle w_z \!-\! w_y, \phi(x+\eta) -  \phi(x) \rangle$, i.e.\ our statistic measures noise-induced feature map weight-difference vector alignment, cf.\ Section~\ref{sec:featurespace}.

\subsection{Log-Odds Robustness}
\vspace{-1mm}
The main idea pursued in this paper is that the robustness properties of the perturbed log-odds statistics are different, dependent on whether $x=x^*$ is naturally generated or whether it is obtained through an (unobserved) adversarial manipulation, $x = x^* + \triangle x$.

Firstly, note that it is indeed very common to use (small-amplitude) noise during training as a way to robustify models or to use regularization techniques which improve model generalization. In our notation this means that for $(x^*,y^*) \sim \datad$, it is a general design goal -- prior to even considering adversarial examples -- that with high probability $f_{y^*,z}(x^*+\eta) \approx f_{y^*,z}(x^*)$, i.e.~that log-odds with regard to the true class remain stable under noise. We generally may expect $f_{y^*,z}(x^*)$ to be negative (favoring the correct class) and slightly increasing under noise, as the classifier may become less certain.

Secondly, we posit that for many existing deep learning architectures, common adversarial attacks find perturbations $\triangle x$ that are \textit{not} robust, but that overfit to specifics of  $x$. We elaborate on this conjecture below by providing empirical evidence and theoretical insights. For the time being, note that \textit{if} this conjecture can be reasonably assumed, then this opens up ways to design statistical tests to identify adversarial examples and even to infer the true class label, which is particularly useful for test time attacks. 

\begin{figure}[t]
\centering
\begin{tabular}{c}
\includegraphics[width=0.83\columnwidth]{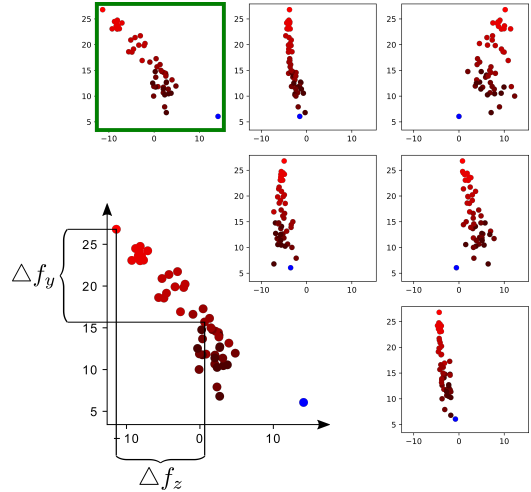}
\vspace{-2mm}
\end{tabular}
\caption{Change of logit scores $f_y$ (on the vertical axis) and $f_z$ (on the horizontal axis) when adding noise to an adversarially perturbed example $x=x^*+\triangle x$. 
Light red dot: $x$. Other red dots: $x+\eta$, with color coding of noise amplitude (light $=$ small, dark $=$ large). Light blue dot: $x^*$. 
Different plots correspond to different candidate classes $z$.
The candidate class in the green box is selected by \eqref{eq:barF} and the plot magnified in the lower left.} 
\label{fig:revert}
\end{figure}

\begin{figure}[t]
\centering
	\adjincludegraphics[width=0.85\linewidth , trim={{0.11\width} {0.1\height} {0.1\width} {0.1\height}},clip]{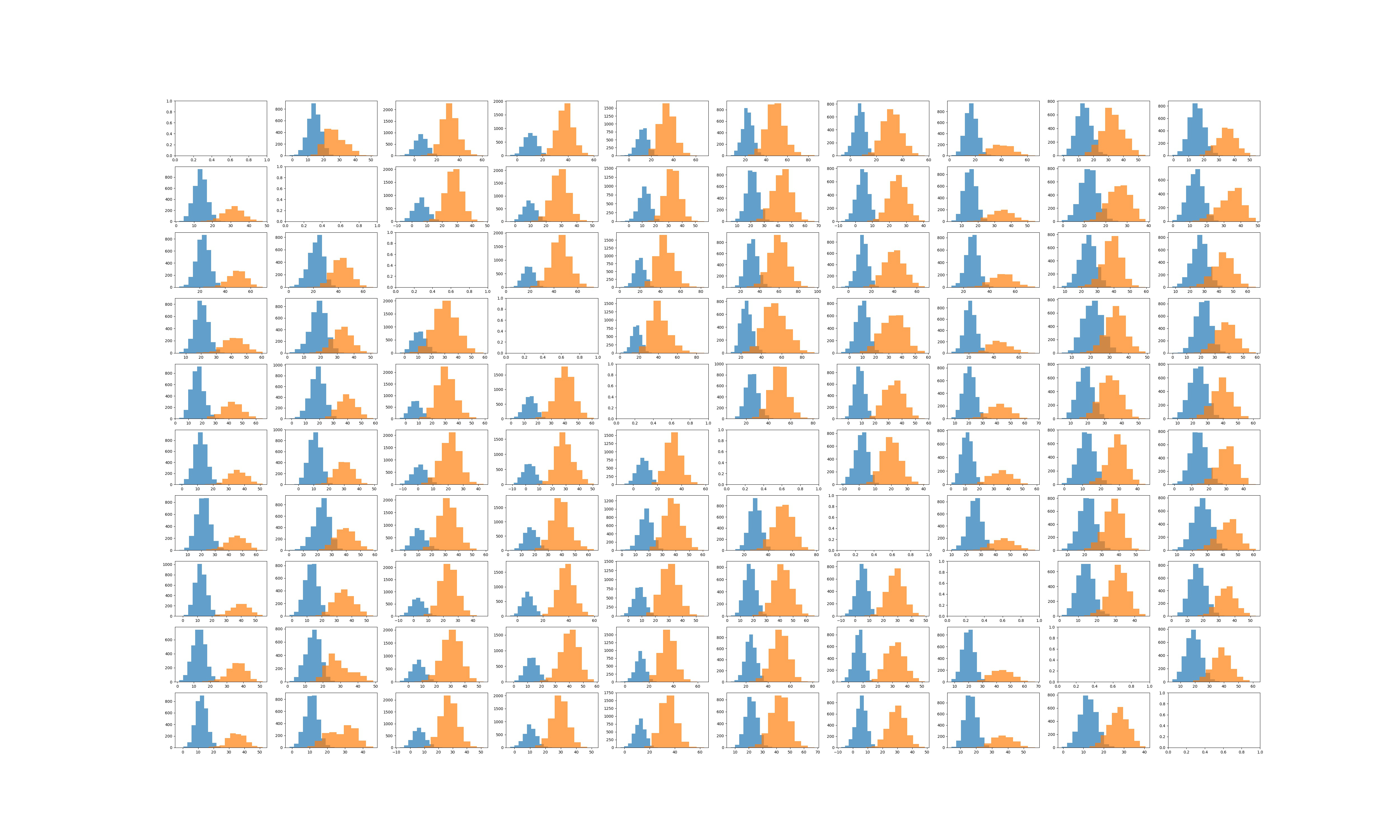}
	\vspace{-2mm}
	\caption{Histograms of the test statistic $\bar g_{y,z}(x)$ aggregated over all data points in the training set. 
	Blue represents natural data, orange represents adversarially perturbed data. Columns correspond to predicted labels $y$, rows to candidate classes $z$.}
\label{fig:weightdifferencehistograms}
\end{figure}

Consider the case of a  test time attack, where we suspect an unknown perturbation $\triangle x$ has been applied such that $F(x^*+\triangle x)=y\neq y^*$. If the perturbation is not robust w.r.t.~the noise process, then this will yield $f_{y,y^*} (x+\eta) > f_{y,y^*} (x) $, meaning that noise will partially undo the effect of the adversarial manipulation and directionally revert the log-odds towards the true class $y^*$ in a way that is statistically captured in the perturbed  log-odds. Figure \ref{fig:revert} (lower left corner) shows this reversion effect. 
Figure~\ref{fig:weightdifferencehistograms} shows an experiment performed on the CIFAR10 data set, which confirms that the histograms of standardized log-odds $\bar g_{y,z}(x)$ (defined below) show a good spearation between clean data $x^*$ and manipulated data points $x= x^*+\triangle x$.

\subsection{Statistical Test \& Corrected Classification}
\label{sec:statisticaltest}

We propose to use the expected perturbed log-odds $\bar g_{y,z}(x) = \mathbf E_\eta \left[\bar g_{y,z}(x, \eta) \right]$ as statistics to test whether $x$ classified as $y$ should be thought of as a manipulated example of (true) class $z$ or not. To that extent, we define thresholds $\tau_{y,z}$, which guarantee a maximal false detection rate (of say 1\%), yet maximize the true positive rate of identifying  adversarial examples. We then flag an example $(x,y:=F(x))$ as (possibly) manipulated, if 
\begin{align}
\max_{z \neq y} \left\{ \bar g_{y,z}(x) - \tau_{y,z} \right\}\ge 0\,,
\label{eq:rejection}
\end{align}
otherwise it is considered clean.

For test time attacks, it may be relevant not only to detect manipulations, but also to \textit{correct} them on the spot. The simplest approach is to define a new classifier $G$ via
\begin{align}
G(x) & = \arg\max_{z} \left\{ \bar g_{y,z}(x) - \tau_{y,z} \right\}, \;\; y := F(x) \,.
\label{eq:barF}
\end{align}
Here we have set $\tau_{y,y}=\bar g_{y,y} = 0$, which sets the correct reference point consistent with \eqref{eq:rejection}.

A more sophisticated approach is to build a second level classifier on top of the perturbed log-odds statistics. We performed experiments with training a logistic regression classifier for each class $y$ on top of the standardized log-odds scores $\bar g_{y,z}(x)$, $y=F(x)$, $z \neq y$. We found this  to further improve classification accuracy, especially in cases where several Z-scores are comparably far above the threshold. See Section~\ref{sec:appendixexperiments} in the Appendix for further details.

% !TEX root = 00_icml2019.tex
\vspace{-1mm}
\section{Feature Space Analysis}
\label{sec:featurespace}

\subsection{Optimal Feature Space Manipulation}

The feature space view allows us to characterize the optimal direction of manipulation for an attack targetting some class $z$. Obviously the log-odds $f_{y^*, z}$ only depend on a single direction in feature space, namely $\triangle w_z = w_z-w_{y^*}$. 

\begin{proposition}
For constraint sets $\mathcal B$ that are closed under orthogonal projections, the optimal attack in feature space takes the form $\triangle \phi^* = \alpha ( w_z - w_{y^*})$ for some $\alpha \ge 0$.
\begin{proof}
\vspace{-2mm}
Assume $\triangle \phi \in \mathcal B$ is optimal. We can decompose $\triangle \phi = \alpha \triangle w_z + v$, where $v \perp \triangle w_z$. $\triangle \phi^*$ achieves the same change in log-odds as $\triangle \phi$ and is also optimal. 
\end{proof}
\end{proposition}

\vspace{-1mm}
\begin{proposition}
If $\triangle \phi$ s.t.~$y = \arg\max_z \langle w_z, \phi(x) + \triangle \phi \rangle$ and $y^* = \arg\max_z \langle w_z, \phi(x) \rangle$, then $\langle \triangle \phi, \triangle w_{y} \rangle \geq 0$. 
\vspace{-2mm}
\begin{proof}
Follows directly from $\phi$-linearity of log-odds.
\end{proof}
\end{proposition}
\vspace{-2mm}
Now, as we treat the deep neural net defining $\phi$ as a black box device, it is difficult to state whether a (near-)optimal feature space attack can be carried out by manipulating in the input space via $\triangle x \mapsto \triangle \phi$. However, we will use some DNN phenomenology as a starting point for making reasonable assumptions that can advance our understanding.

\subsection{Pre-Image Problems} 

The feature space view suggests to search for a pre-image of the optimal manipulation  $\phi(x) + \triangle \phi^*$ or at least a manipulation $\triangle x$  such that $\| \phi(x)+\triangle \phi^* - \phi(x+\triangle x) \|^2$ is small. 
A na\"ive approach would be to linearize $\phi$ at $x$ and use the Jacobian,
\begin{align} 
\phi(x + \triangle x) = \phi(x) + J_\phi(x) \triangle x + O(\|\triangle x\|^2)\,.
\end{align}
Iterative improvements could then be obtained by inverting (or pseudo-inverting) $J_\phi(x)$, but are known to be plagued by instabilities. 
A popular alternative is the so-called Jacobian transpose method from inverse kinematics \cite{buss2004introduction, wolovich1984computational, balestrino1984robust}. 
This can be motivated by a simple observation 
\begin{proposition}
Given an input $x$ as well as a target direction $\triangle \phi$ in feature space. Define $\triangle x := J_\phi^\top(x) \triangle \phi$ and assume that $\langle J_\phi\triangle x, \triangle \phi \rangle >0$. Then there exists an $\epsilon >0$ (small enough) such that $x^+ := x + \epsilon \triangle x$ is a better pre-image in that $\| \phi(x) + \triangle \phi - \phi(x^+)\| < \| \triangle \phi\|$.
\vspace{-2mm} 
\begin{proof} Follows from Taylor expansion of $\phi$. \vspace{-1mm} \end{proof}
\end{proposition}
It turns out that by the chain rule, we get for any loss $\ell$ defined in terms of features $\phi$,
\begin{align}
\nabla_x (\ell \circ \phi)(x) = J_\phi^\top(x)  \nabla_\phi \ell(\phi) |_{\phi=\phi(x)}\,.
\end{align}
With the soft-max loss $\ell(x) = -f_y(x) + \log \sum_z \exp[f_z(x)]$ and in case of $f_{y^*}(x) \gg f_z(x)$ one gets
\begin{align}
\nabla_\phi \ell(\phi) = w_{y^*} - w_y = - \triangle w_y\,.
\end{align}
This shows that a gradient-based iterative attack is closely related to solving the pre-image problem for finding an optimal feature perturbation  via the Jacobian transpose method.

\subsection{Approximate Rays and Adversarial Cones} 

If an adversary had direct control over the feature space representation, optimal attack vectors could always be found along the ray $\triangle w_z$. 
As the adversary has to work in input space, this may only be possible in approximation however. 
Experimentally, we have found that an optimal perturbation typically defines a ray in input space, $x + t \triangle x$ ($t \ge 0$), yielding a feature-space trajectory $\phi(t) = \phi(x + t \triangle x)$ for which the rate of change along $\triangle w_z$ is nearly constant over a relevant range of $t$, see Figures \ref{fig:featurespacedistalignment}\, \& \,\ref{fig:softmaxpredictions}. 
While the existence of such rays obviously plays in the hand of an adversary, it remains an open theoretical question to eluciate properties of the model architecture causing such vulnerabilities.  

As adversarial directions are expected to be suscpetible to angular variations (otherwise they would be simple to find and pointing at a general lack of model \textit{robustness}), we conjecture that geometrically optimal adversarial manipulations are embedded in a cone-like structure, which we call \textit{adversarial cone}. Experimental evidence for the existence of such cones is visualized  in Figure \ref{fig:adversarialcone}. It is a virtue of the commutativity of applying the adversarial $\triangle x$ and random noise $\eta$ that our statistical test can reliably detect such adversarial cones.

% !TEX root = 00_icml2019.tex

\section{Experimental Results}
\label{sec:experiments}

\begin{table}[t]
\centering
	\caption{Baseline test set accuracies on clean and PGD-perturbed examples for the models we considered.}\label{tbl:accuracies}
	\vskip 0.15in
	\scriptsize
	\begin{sc}
	\begin{tabular}{llr}
		\toprule
		Dataset & Model & Test set accuracy \\
		& & (clean / pgd) \\
		\midrule
		CIFAR10 & WResNet & 96.2\% / 2.60\% \\
		& CNN7 & 93.8\% / 3.91\% \\
		& CNN4 & 73.5\% / 14.5\% \\
		\\[-2mm]
		ImageNet & Inception V3 & 76.5\% / 7.2\% \\
		& ResNet 101 & 77.0\% / 7.2\% \\
		& ResNet 18 & 69.2\% / 6.5\% \\
		& VGG11(+BN) & 70.1\% / 5.7\% \\
		& VGG16(+BN) & 73.3\% / 6.1\% \\
		\bottomrule
	\end{tabular}
\end{sc}
\end{table}

\subsection{Datasets, Architectures \& Training Methods} 
In this section, we provide experimental support for our theoretical propositions and we benchmark our detection and correction methods on various architectures of deep neural networks trained on the CIFAR10 and ImageNet datasets.
For CIFAR10, we compare the WideResNet implementation of \citet{madry2017towards}, a 7-layer CNN with batch normalization and a vanilla 4-layer CNN. 
In the following, if nothing else is specified, we use the 7-layer CNN as a default platform, since it has good test set accuracy at relatively low computational requirements. 
For ImageNet, we use a selection of models from the torchvision package \cite{marcel2010torchvision}, including Inception V3, ResNet101 and VGG16.
Further details can be found in the Appendix.

As a default attack strategy we use an $L^\infty$-norm constrained PGD white-box attack. 
The attack budget $\epsilon_\infty$ was chosen to be the smallest value such that almost all examples are successfully attacked. 
For CIFAR10 this is $\epsilon_\infty=8/255$, for ImageNet $\epsilon_\infty=2/255$. 
We experimented with a number of different PGD iterations and found that the detection rate and corrected classification accuracy are nearly constant across the entire range from 10 up to 1000 attack iterations, as shown in Figure~\ref{fig:attackiterations} in the Appendix. For the remainder of this paper, we thus fixed the number of attack iterations to  20.
Table~\ref{tbl:accuracies} shows test set accuracies for all considered models on both clean and adversarial samples.

We note that the detection test in \Eqref{eq:rejection} as well as the basic correction algorithm in \Eqref{eq:barF} are completely attack agnostic.
The only stage that explicitly includes an adversarial attack model is the second-level logistic classifier based correction algorithm, which is trained on adversarially perturbed samples, as explained in Section~\ref{sec:appendixexperiments} in the Appendix. 
While this could in principle lead to overfitting to the particular attacks considered, we empirically show that the second-level classifier based correction algorithm performs well under attacks not seen during training, cf.\ Section~\ref{subsection:unseen}, as well as specifically designed counter-attacks, cf.\ Section~\ref{sec:counterattack}.

\begin{figure}[t]
\begin{tabular}{@{}l@{}c@{}l@{}c@{}} 
	\begin{turn}{90}\hspace{28pt} {$\scriptscriptstyle || \triangle\phi ||_2$} \end{turn}\hspace{1pt} & \adjincludegraphics[width=0.46\linewidth , trim={{0.055\width} {0.065\height} {0.04\width} {0.067\height}},clip]{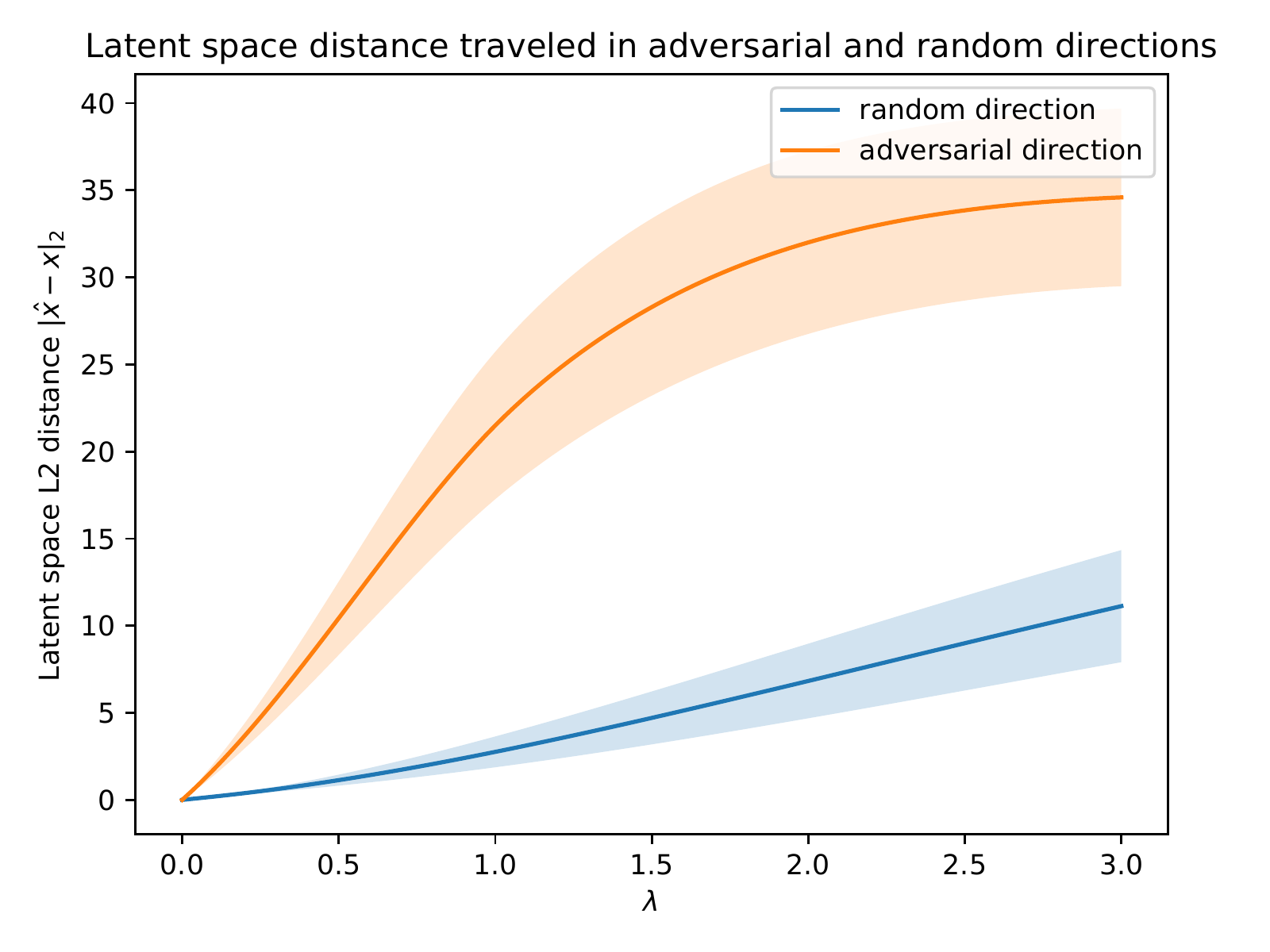} &
	\hspace{5pt}\begin{turn}{90}\hspace{25pt} {$\scriptscriptstyle\langle \triangle\phi, \triangle w \rangle$} \end{turn} & \adjincludegraphics[width=0.46\linewidth , trim={{0.055\width} {0.065\height} {0.038\width} {0.067\height}},clip]{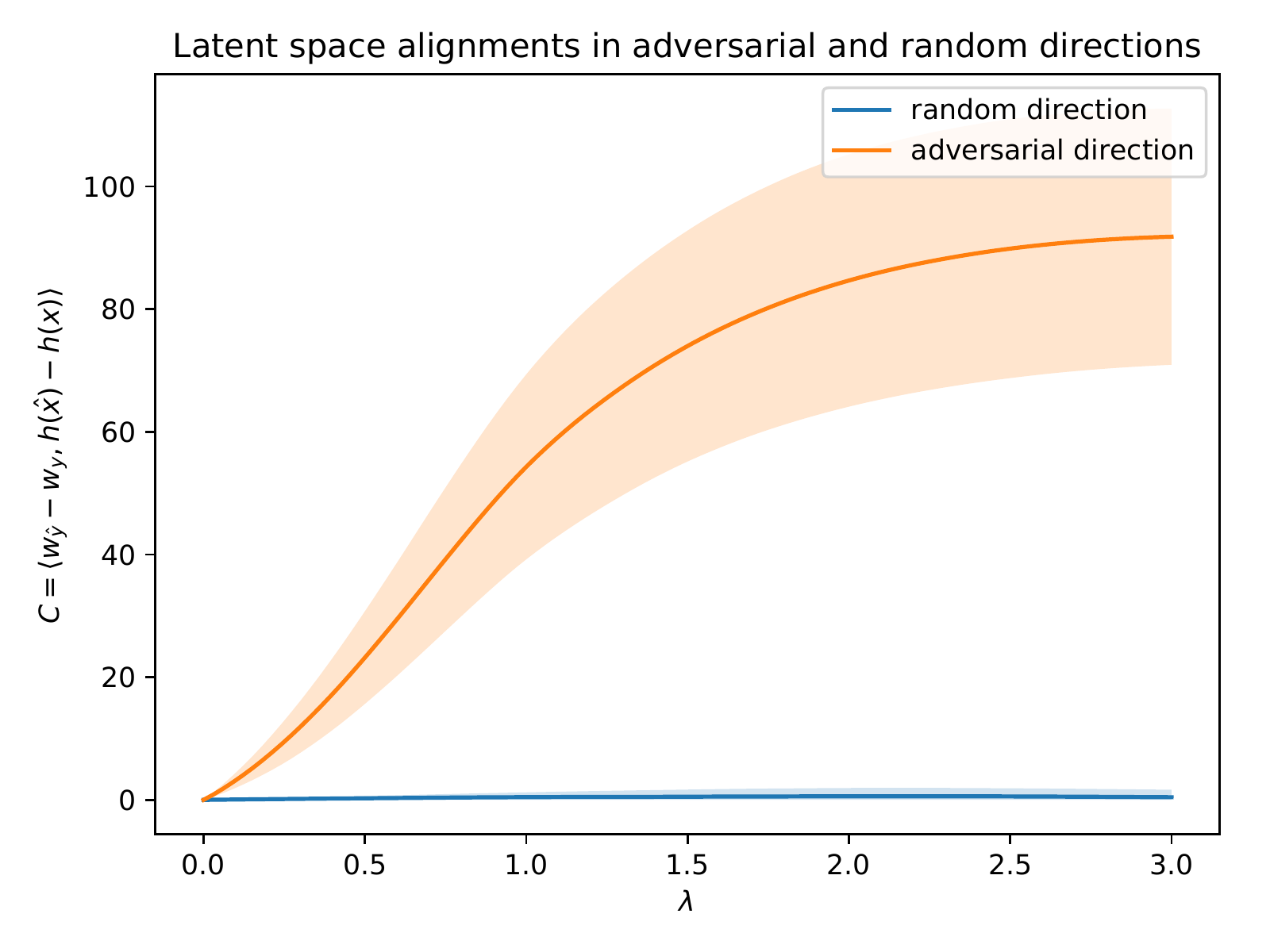}  \\[-5pt]
	& $\scriptstyle t$ & & $\scriptstyle t$ \\[-5pt]
\end{tabular}
\caption{(Left) Norm of the induced feature space perturbation along adversarial and random directions.
(Right) Weight-difference alignment. For the adversarial direction, the alignment with the weight-difference between the true and adversarial class is shown.
For the random direction, the largest alignment with any weight-difference vector is shown. See also Figure~\ref{fig:robustnesstonoise}. }
\label{fig:featurespacedistalignment}
\vspace{-2mm}
\end{figure}

%---------------------------------------------------------------------------------------------------------------------------%
% DETECTABILITY of ADVERSARIAL EXAMPLES
%---------------------------------------------------------------------------------------------------------------------------%
\subsection{Detectability of Adversarial Examples}
\label{sec:AmbientSpace}

{\bf Induced feature space perturbations.}
We compare the norm of the induced feature space perturbation $|| \triangle\phi ||_2$ 
along adversarial directions $x^*\!+\!t\triangle x$ with that along random directions $x^*\!+\!t\eta$ (where the expected norm of the noise is set to be approximately equal to the expected norm of the adversarial perturbation).  
We also compute the alignment $\langle \triangle\phi, \triangle w \rangle$ between the induced feature space perturbation and certain weight-difference vectors: 
For the adversarial direction, the alignment is computed w.r.t.\ the weight-difference vector between the true and adversarial class, 
for the random direction, the largest alignment with any weight-difference vector is computed.

The results are reported in Figure~\ref{fig:featurespacedistalignment}. 
The plot on the left shows that iterative adversarial attacks induce feature space perturbations that are significantly larger than those induced by random noise.
The plot on the right shows that the attack-induced weight-difference alignment
is significantly larger than the noise-induced one. 
The plot on the right in Figure~\ref{fig:robustnesstonoise} in the Appendix further shows that the noise-induced weight-difference alignment is significantly larger for the adversarial example than for the natural one.
Combined, this indicates that \textit{adversarial examples cause atypically large feature space perturbations along the weight-difference direction $\triangle w_y = w_y-w_{y^*}$, with $y = F(x)$.}

{\bf Distance to decision boundary.}
Next, we investigate how the distance to the decision boundary for adversarial examples compares with that of their unperturbed counterpart.
To this end, we measure the logit cross-over when linearly interpolating between an adversarially perturbed example and its natural counterpart, 
i.e.\ we measure $t \in [0, 1]$ s.t.\ $f_{y^*}(x^* \!+\! t\triangle x) \simeq f_{y}(x^* \!+\! t\triangle x)$, where $y = F(x^* \!+\! \triangle x)$. 
We also measure the average $L^2$-norm of the DeepFool perturbation $\triangle x(t)$, required to cross the nearest decision boundary\footnote{The DeepFool attack aims to find the shortest path to the nearest decision boundary. We additionally augment DeepFool by a binary search to hit the decision boundary precisely.}, for all interpolants $x(t) = x^* \!+\! t\triangle x$.

\begin{figure}[t]
\centering
\begin{tabular}{@{}l@{\hspace{4pt}}c@{}} 
	\begin{turn}{90}\hspace{33pt} {\tiny distance to db} \end{turn} & \adjincludegraphics[width=0.57\linewidth , trim={{0.07\width} {0.06\height} {0.1\width} {0.11\height}},clip]{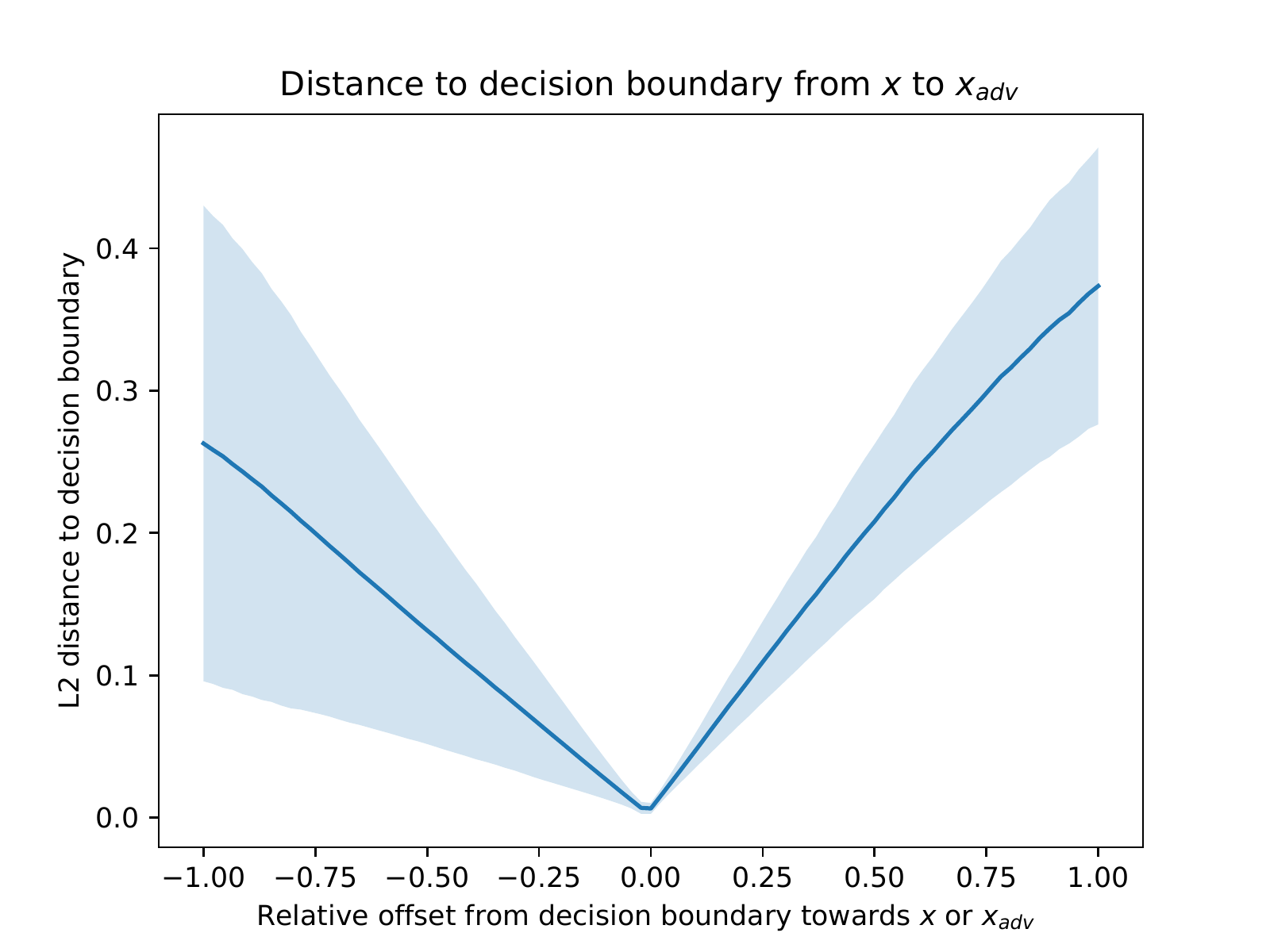} \\[-3pt]
	& \hspace{7pt}{\tiny relative offset to logit cross-over point} \\[-2pt]
\end{tabular}
\caption{Average distance to the decision boundary when interpolating from natural examples to adversarial examples. 
The horizontal axis shows the relative offset of the interpolant~$x(t)$ to the logit cross-over point located at the origin $0$. 
For each interpolant~$x(t)$, the distance to the nearest decision boundary is computed. 
The plot shows that natural examples are slightly closer to the decision boundary than adversarial examples.}
\label{fig:disttodb}
\end{figure}

\begin{figure}
\begin{center}
	\includegraphics[width=0.46\linewidth]{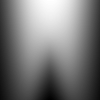}
\end{center}
\vspace{-2mm}
\caption{Adversarial cone. 
The plot shows the averaged softmax prediction for the natural class over ambient space hyperplanes spanned by the adversarial perturbation (on the vertical axis) and randomly sampled orthogonal vectors (on the horizontal axis).
The natural sample is located one-third from the top, the adversarial sample one third from the bottom on the vertical axis through the middle of the plot.}
\label{fig:adversarialcone}
\end{figure}

We find that the mean logit cross-overs is at $\bar{t}=0.43$. 
Similarly, as shown in Figure~\ref{fig:disttodb}, the mean $L^2$-distance to the nearest decision boundary is $0.37$ for adversarial examples, compared to $0.27$ for natural ones. 
Hence, natural examples are even slightly closer to the decision boundary.
\textit{We can thus rule out the possibility that adversarial examples are detectable because of a trivial discrepancy in distance to the decision boundary}.

{\bf Neighborhood of adversarial examples.}
We measure the ratio of the `distance between the adversarial and the corresponding unperturbed example' to the `distance between the adversarial example and the nearest other neighbor (in either training or test set)', i.e.\ we compute 
$||x - x^*||_2 / ||x - x^{\rm nn}||_2$ over a number of samples in the test set, 
for various $L^\infty$- \& $L^2$-bounded PGD attacks (with $\epsilon_2 = \sqrt{D}\epsilon_\infty$).

We consistently find that the ratio is sharply peaked around a value much smaller than one. 
E.g.\ for $L^\infty$-PGD attack with $\epsilon_\infty\!=\!8/255$ we get $0.075 \!\pm\! 0.018$,
while for the corresponding $L^2$-PGD attack we obtain $0.088 \!\pm\! 0.019$.
Further values can be found in Table~\ref{table:ratioofdistances} in the Appendix. 
We note that similar findings have been reported by \mbox{\citet{tramer2017space}}.
Hence, ``perceptually similar'' \emph{adversarial examples are much closer to the unperturbed sample than to any other neighbor in the training or test set}.

We would therefore naturally expect that
the feature representation is more likely to be shifted to the original unperturbed class rather than any other neighboring class 
when the adversarial example is convolved with random noise.

To investigate this further, we plot the softmax predictions when adding noise to the adversarial example.
The results are reported in Figure~\ref{fig:softmaxpredictions} in the Appendix.
The plot on the left shows that the probability of the natural class increases faster than the probability of the highest other class when adding noise with a small to intermediate magnitude to the adversarial example. 
However, the probability of the natural class never climbs to be the highest probability of all classes, which is why simple addition of noise to an adversarial example does not recover the natural class in general.

{\bf Adversarial Cones.} To visualize the ambient space neighborhood around natural and adversarially perturbed samples, 
we plot the averaged classifier prediction for the natural class over hyperplanes spanned by the adversarial perturbation and randomly sampled orthogonal vectors,
i.e.\ we plot $\E_{n} [ F_{y^*}(x^* + t\triangle x + s n ) ]$ for $s \in [-1,1], t \in [-1, 2]$
with $s$ along the horizontal and $t$ along the vertical axis, 
where $F_{y^*}$ denotes the softmax 
and $\E_{n}$ denotes expectation over random vectors $n \perp \triangle x$ with approximately equal norm.

Interestingly, the plot reveals that adversarial examples are embedded in a cone-like structure, i.e.\  the adversarial sample is statistically speaking ``surrounded'' by the natural class, as can be seen from the gray rays confining the adversarial cone. 
This confirms our theoretical argument that the noise-induced feature variation tends to have a direction that is indicative of the natural class when the input is adversarially perturbed. 

It is a virtue of the commutativity of applying the adversarial and random noise, i.e.\ $x^* + \triangle x +\eta$ vs.\ $x^* + \eta + \triangle x$, that our method can reliably detect such adversarial cones.

%---------------------------------------------------------------------------------------------------------------------------%
% DETECTION RATES
%---------------------------------------------------------------------------------------------------------------------------%
\subsection{Detection rates and classification accuracies} 
\label{sec:detectionrates}

In the remainder of this section we present the results of various performance evaluations. 
The reported detection rates measure how often our method classifies a sample as being adversarial, corresponding to the False Positive Rate if the sample is clean and to the True Positive Rate if it was perturbed. 
We also report accuracies for the predictions made by the logistic classifier based correction method.

Tables~\ref{tbl:aftermethod:detection} and~\ref{tbl:aftermethod:accuracy} report the detection rates of our statistical test and accuracies of the corrected predictions. 
Our method manages to detect nearly all adversarial samples, seemingly getting better as models become more complex, while the false positive rate stays around $1\%$. 
Further\footnote{Due to computational constraints, we focus on the CIFAR10 models in the remainder of this paper.}, 
our second-level logistic-classifier based correction method manages to reclassify almost all of the detected adversarial samples to their respective source class successfully, resulting in test set accuracies on adversarial samples within $5\%$ of the respective test set accuracies on clean samples. Also note that due to the low false positive rate, the drop in performance on clean samples is negligible.

\begin{table}
\vspace{-2mm}
\centering
	\caption{Detection rates of our statistical test.} \label{tbl:aftermethod:detection}
	\vskip 0.15in
	\scriptsize
	\begin{sc}
	\begin{tabular}{llr}
		\toprule
		Dataset & Model & Detection rate \\
		& &(clean / pgd) \\
		\midrule
		CIFAR10 &WResNet & 0.2\% / 99.1\% \\
		&CNN7 & 0.8\% / 95.0\% \\
		&CNN4 & 1.4\% / 93.8\% \\
		\\[-2mm]
		ImageNet&Inception V3 & 1.9\% / 99.6\%\\
		&ResNet 101 & 0.8\% / 99.8\% \\
		&ResNet 18 & 0.6\% / 99.8\% \\
		&VGG11(+BN) & 0.5\% / 99.9\% \\
		&VGG16(+BN) & 0.3\% / 99.9\%\\
		\bottomrule
	\end{tabular}
\end{sc}
\vspace{-3mm}
\end{table}
\begin{table}
\centering
	\caption{Accuracies of our correction method.} \label{tbl:aftermethod:accuracy}
	\vskip 0.15in
	\scriptsize
	\begin{sc}
	\begin{tabular}{llr}
		\toprule
		Dataset & Model & Accuracy \\
		& & (clean / pgd) \\
		\midrule
		CIFAR10 &WResNet &  96.0\% / 92.7\% \\
		&CNN7 & 93.6\% / 89.5\% \\
		&CNN4 & 71.0\% / 67.6\% \\
		\bottomrule
	\end{tabular}
\end{sc}
\end{table}

\subsection{Effective strength of adversarial perturbations.} 
We measure how the detection rate and reclassification accuracy of our method depend on the effective attack strength.
To this end, we define the effective Bernoulli-$q$ strength of $\epsilon$-bounded adversarial perturbations as the attack success rate when each entry of the perturbation $\triangle x$ is individually accepted with probability $q$ and set to zero with probability $1-q$. 
For $q=1$ we obtain the usual adversarial misclassification rate.
We naturally expect weaker attacks to be less effective but also harder to detect than stronger ones.

\begin{figure}
\centering
	\begin{tabular}{@{}l@{}c@{\hspace{5pt}}l@{}c@{}} 
	\begin{turn}{90}\hspace{9pt} {\tiny detection rate / accuracy} \end{turn} & \adjincludegraphics[width=0.48\linewidth , trim={{0.05\width} {0.085\height} {0.02\width} {0.073\height}},clip]{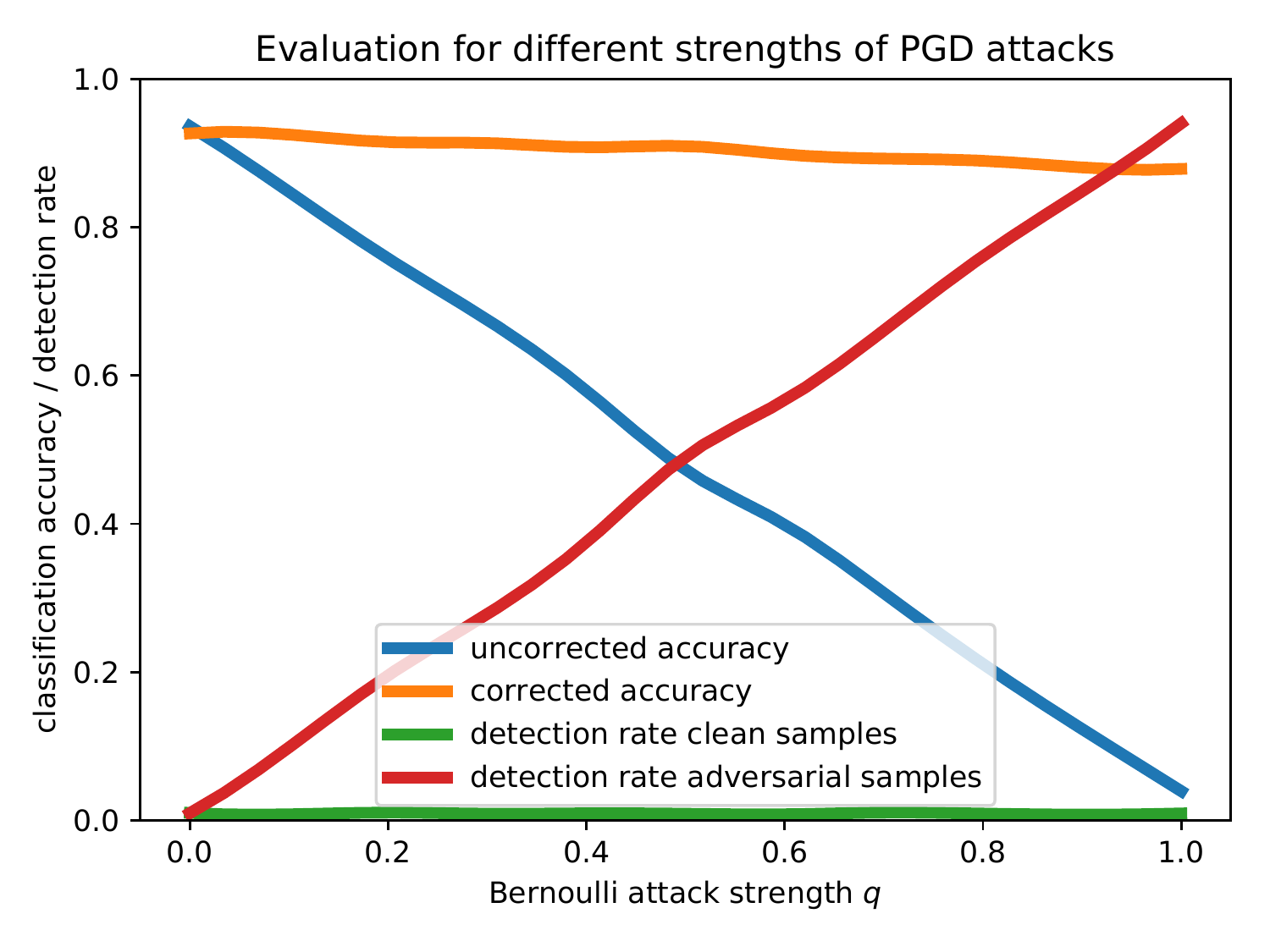} &
	\begin{turn}{90}\hspace{9pt} {\tiny detection rate / accuracy} \end{turn} & \adjincludegraphics[width=0.48\linewidth , trim={{0.05\width} {0.085\height} {0.02\width} {0.073\height}},clip]{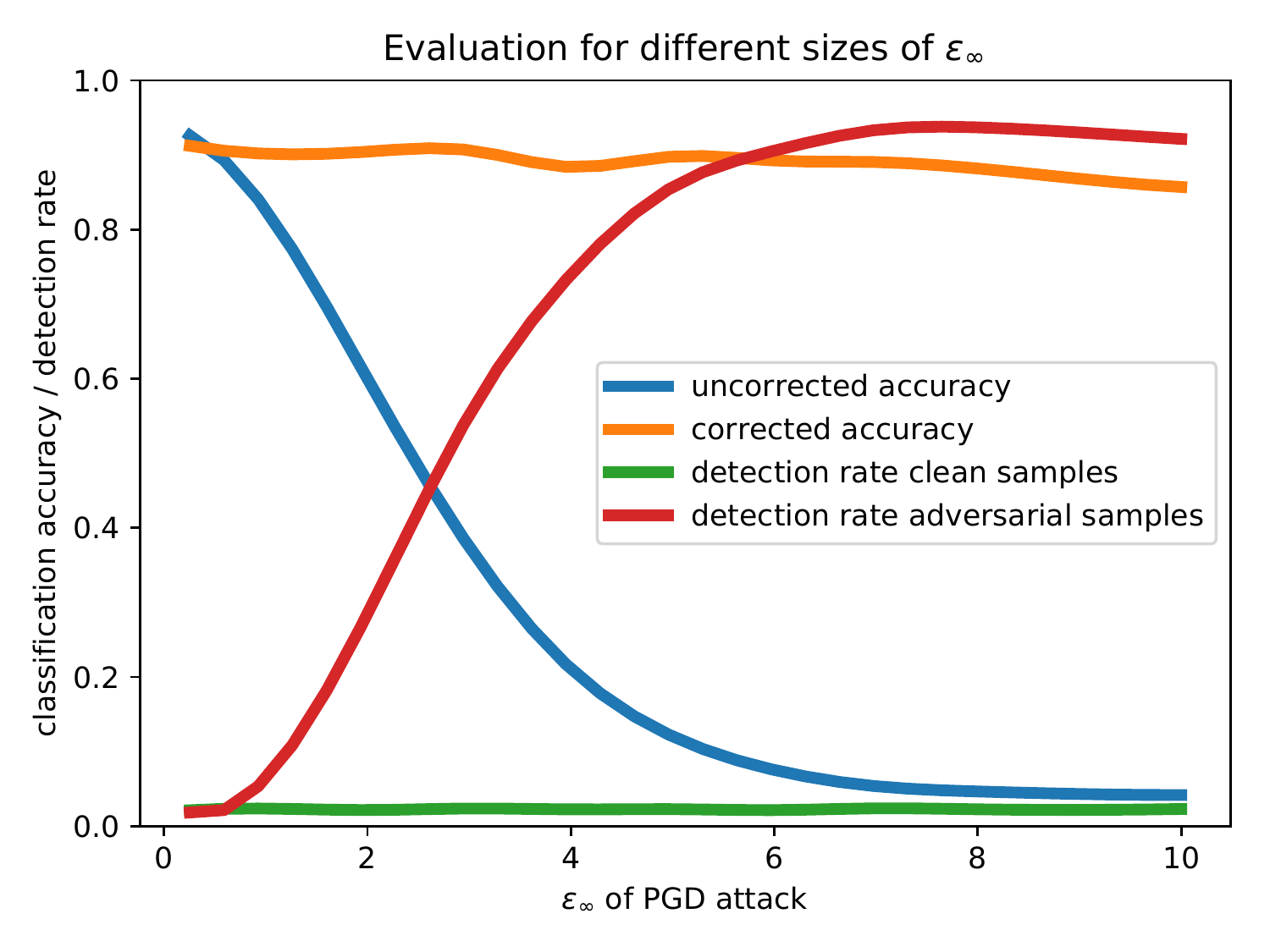} \\[-5pt]
	& \,\,{\tiny Bernoulli attack strength q} & & $\scriptscriptstyle\epsilon_\infty$ \\[-5pt]
	\end{tabular}
	\caption{Detection rate and reclassification accuracy as a function of the effective attack strength.
	The uncorrected classifier accuracy decreases as the attack strength increases, both in terms of the attack budget $\epsilon_\infty$ as well as in terms of the fraction $q$ of accepted perturbation entries.
	Meanwhile, the detection rate of our method increases at such a rate that the corrected classifier manages to compensate for the decay in uncorrected accuracy.}
	\label{fig:attackstrength}
\end{figure}

The results are reported in Figure \ref{fig:attackstrength}.
We can see that the \mbox{uncorrected} accuracy of the classifier decreases monotonically as the effective attack strength increases, both in terms of the attack budget $\epsilon_\infty$ as well as in term of the fraction $q$ of accepted perturbation entries.
Meanwhile, the detection rate of our method increases at such a rate that the corrected classifier manages to compensate for the decay in uncorrected accuracy, due to the decrease in effective strength of the perturbations, across the entire range considered.

\subsection{Comparison to Adversarial Training} 
For comparison, we also report test set and white-box attack accuracies for adversarially trained models.
{\citet{madry2017towards}'s \mbox{WResNet} was available as an adversarially pretrained variant, while the other models were adversarially trained as outlined in Section~\ref{sec:appendixexperiments} in the Appendix.
The results for the best performing classifiers are shown in Table~\ref{tbl:advtraining}.
We can see that the accuracy on adversarial samples is significantly lower while the drop in performance on clean samples is considerably larger for adversarially trained models compared to our method.

\begin{table}
\vspace{-2mm}
\centering
	\caption{Test set accuracies for adversarially trained models.} \label{tbl:advtraining}
	\vskip 0.15in
	\scriptsize
	\begin{sc}
	\begin{tabular}{llr}
		\toprule
        Dataset & Adversarially & Accuracy \\
                & trained model & (clean / pgd) \\
		\midrule
        CIFAR10 & WResNet & 87.3\% / 55.2\% \\
		        & CNN7 & 82.2\% / 44.4\% \\
		        & CNN4 &68.2\% / 40.4\% \\
		\bottomrule
	\end{tabular}
\end{sc}
\end{table}

\subsection{Defending against unseen attacks}
\label{subsection:unseen}
Next, we evaluate our method on adversarial examples created by attacks that are different from the $L_\infty$-constrained PGD attack used to train the second-level logistic classifier.
The rationale is that the log-odds statistics of the unseen attacks could be different from the ones used to train the logistic classifier. We thus want to test whether it is possible to evade correct reclassification by switching to a different attack. 
As alternative attacks we use an $L^2$-constrained PGD attack as well as the $L^2$-Carlini-Wagner attack.

The baseline accuracy of the undefended CNN7 on adversarial examples is $4.8\%$ for the $L^2$-PGD attack and $3.9\%$ for the Carlini-Wagner attack.
Table~\ref{tbl:advtraining:defended} shows the detection rates and corrected classification accuracies of our method.
As can be seen, there is only a slight decrease in performance, i.e.\ our method remains capable of detecting and correcting most adversarial examples of the previously unseen attacks.

\begin{table}
\vspace{-2mm}
	\centering
		\caption{CIFAR10 detection rates and reclassification accuracies on adversarial samples from attacks that have not been used to train the second-level logistic classifier.} \label{tbl:advtraining:defended}
		\vskip 0.15in
		\scriptsize
		\begin{sc}
	\begin{tabular}{lrr}
		\toprule
		Attack & Detection rate & Accuracy \\
		& (clean / attack) & (clean / attack) \\
		\midrule
		$L^2$-PGD & 1.0\% / 96.1\% & 93.3\% / 92.9\% \\
		$L^2$- CW & 4.8\% / 91.6\% & 89.7\% / 77.9\% \\
		\bottomrule
	\end{tabular}
\end{sc}
\end{table}

\subsection{Defending against defense-aware attacks}
\label{sec:counterattack}
Finally, we evaluate our method in a setting where the attacker is fully aware of the defense, in order to see if the defended network is susceptible to cleverly designed counter-attacks.
Since our defense is built on random sampling from noise sources that are under our control, the attacker will want to craft perturbations that perform well \emph{in expectation} under this noise. 
The optimality of this strategy in the face of randomization-based defenses was established in \citet{carlini2017adversarial} (cf.\ their recipe to attack the dropout randomization defense of \citet{feinman2017detecting}).  
Specifically, each PGD perturbation is computed for a loss function that is an empirical average over $K=100$ noise-convolved data points, with the same noise source as used for detection.
(We have also experimented with other variants such as backpropagating through our statistical test and found the above approach by \citet{carlini2017adversarial} to work best.)

The undefended accuracies under this attack for the models under consideration are:  WResNet $2.8\%$, CNN7 $3.6\%$ and CNN4 $14.5\%$.
Table~\ref{tbl:mean:defended} shows the corresponding detection rates and accuracies after defending with our method. 
Compared to Section~\ref{subsection:unseen}, the drop in performance is larger, as we would expect for a defense-aware counter-attack, 
however, both the detection rates and the corrected accuracies remain remarkably high compared to the undefended network.

\begin{table}
\vspace{-2mm}
\centering
	\caption{CIFAR10 detection rates and reclassification accuracies on clean and adversarial samples from the defense-aware attacker.} \label{tbl:mean:defended}
	\vskip 0.15in
	\scriptsize
	\begin{sc}
	\begin{tabular}{lrr}
		\toprule
		Model & Detection rate & Accuracy \\
		& (clean / attack) & (clean / attack) \\
		\midrule
		WResNet & 4.5\% / 71.4\% & 91.7\% / 56.0\% \\
		CNN7 & 2.8\% / 75.5\%  & 91.2\% / 56.6\% \\
		CNN4 & 4.1\% / 81.3\% & 69.0\% / 56.5\% \\
		\bottomrule
	\end{tabular}
\end{sc}
\end{table}

\subsection{Comparison with related detection methods}
\label{sec:sota}

In this last section we provide a quantitative comparison with two of the leading detection methods: feature squeezing of \citet{xu2017feature} and dropout randomization (aka Bayesian neural network uncertainty) of \citet{feinman2017detecting}.
The reason we compare against those two is that \citet{carlini2017adversarial} consider dropout randomization to be the only defense, among the ten methods they surveyed (including the other two detection methods we mentioned in more detail in the related work section), that is not completely broken, while the more recent feature squeezing method was selected because it was evaluated extensively on comparable settings to ours.

On CIFAR10, feature squeezing\footnote{We report their best joint detection ensemble of squeezers.} (DenseNet)
significantly enhances the model robustness against $L^2$-CW attacks, while it is considerably less effective against PGD attacks, which the authors suspect could due to feature squeezing being better suited to mitigating smaller perturbations.
For $L^2$-CW attacks, they report a detection rate of $100\%$ (FPR $<5\%$) and corrected accuracies of $89\%$ on clean and $83\%$ on adversarial examples, which is  slightly better than our numbers in Table~\ref{tbl:advtraining:defended}. We would like to note however that we calibrated our method on $L^\infty$-constrained perturbations and that our numbers could probably be improved by calibrating on $L^2$-constrained perturbations instead. 
For $L^\infty$-PGD, feature squeezing achieves a detection rate of $55\%$ (FPR $<5\%$), with corrected accuracies of $89\%$ on clean and $56\%$ on adversarial examples,
whereas our method achieves a detection rate of $99\%$ (FPR $<1\%$), with accuracies of $96\%$ on clean and $92\%$ on adversarial samples respectively. 
    On ImageNet, feature squeezing (MobileNet) achieves a detection rate of $64\%$ (FPR $5\%$) for $L^\infty$-PGD, while our method achieves a detection rate of $99\%$ (FPR $1\%$).
\citet{xu2017feature} only evaluate feature squeezing against defense-aware attacks on MNIST, finding that their method is not immune.

\citet{feinman2017detecting} do not report individual true and false detection rates. They do however show the ROC curve and report its AUC:
compare their BIM-B curve (PGD with fixed number of iterations) in Figures 9c \& 10 with our Figure~\ref{fig:roc} in the Appendix.
On CIFAR10 (ResNet) \citet{carlini2017adversarial} were able to fool the dropout defense with $98\%$ success, i.e.\ the detection rate is $2\%$ for the defense-aware $L^2$-CW attack.
Our method achieves a detection rate of $71.4\%$ (FPR $4.5\%$) in a comparable setting.

% !TEX root = 00_icml2019.tex

\section{Conclusion}
\label{sec:conclusion}

We have shown that adversarial examples exist in cone-like regions in very specific directions from their corresponding natural examples.
Based on this, we design a statistical test of a given sample's log-odds robustness to noise that can predict with high accuracy if the sample is natural or adversarial and recover its original class label, if necessary.
Further research into the properties of network architectures is necessary to explain the underlying cause of this phenomenon.
It remains an open question which current model families follow this paradigm and whether criteria exist which can certify that a given model is immunizable via our method.

\section*{Acknowledgements}
We would like to thank Sebastian Nowozin, Aurelien Lucchi, Michael Tschannen, Gary Becigneul, Jonas Kohler and the dalab team for insightful discussions and helpful comments.

\bibliography{bibliography}
\bibliographystyle{icml2019}

% !TEX root = 00_icml2019.tex
\clearpage
\section{Appendix}

\subsection{Experiments.}
\label{sec:appendixexperiments}

Further details regarding the implementation:

{\bf Details on the models used.}
All models on ImageNet are taken as pretrained versions from the torchvision\footnote{https://github.com/pytorch/vision} python package.
For CIFAR10, both CNN7\footnote{https://github.com/aaron-xichen/pytorch-playground} as well as WResNet\footnote{https://github.com/MadryLab/cifar10\_challenge} are available on GitHub as pretrained versions.
The CNN4 model is a standard convolutional network with layers of 32, 32, 64 and 64 channels, each using $3\times 3$ filters and each layer being followed by a ReLU nonlinearity and $2\times 2$ MaxPooling.
The final layer is fully connected.

{\bf Training procedures.}
We used pretrained versions of all models except CNN4, which we trained for 50 epochs with RMSProp and a learning rate of 0.0001.
For adversarial training, the models were trained for 50, 100 and 150 epochs using mixed batches of clean and corresponding adversarial (PGD) samples, matching the respective training schedule and optimizer settings of the clean models. We report results for the best performing variant.
The exception to this is the WResNet model, for which an adversarially trained version was already available.

{\bf Setting the thresholds.}
The thresholds $\tau_{y, z}$ are set such that our statistical test achieves the highest possible detection rate (aka True Positive Rate) 
at a prespecified False Positive Rate of less than $1\%$ ($5\%$ for Sections~\ref{subsection:unseen} and \ref{sec:counterattack}), computed on a hold-out set of natural and adversarially perturbed samples.

{\bf Determining attack strengths.}
For the adversarial attacks we consider, we can choose multiple parameters to influence the strength of the attack. 
Usually, as attack strength increases, at some point there is a sharp increase in the fraction of samples in the dataset where the attack is successful. 
We chose our attack strength such that it is the lowest value after this increase, which means that it is the lowest value such that the attack is able to successfully attack most of the datapoints. 
Note that weaker attacks generate adversarial samples that are closer to the original samples, which makes them harder to detect than excessively strong attacks.

{\bf Noise sources.} Adding noise provides a non-atomic view, probing the classifiers output in an entire neighborhood around the input.
In practice we sample noise from a mixture of different sources: Uniform, Bernoulli and Gaussian noise with different magnitudes.
The magnitudes are sampled from a log-scale.
For each noise source and magnitude, we draw 256 samples as a base for noisy versions of the incoming datapoints, although we have not observed a large drop in performance using only the single best combination of noise source and magnitude and using less samples, which speeds up the wall time used to classify a single sample by an order of magnitude.
For detection, we test the sample in question against the distribution of each noise source, then we take a majority vote as to whether the sample should be classified as adversarial.

{\bf Plots.}
All plots containing shaded areas have been repeated over the dataset.
In these plots, the line indicates the mean measurement and the shaded area represents one standard deviation around the mean.

{\bf Wall time performance.}
Since for each incoming sample at test time, we have to forward propagate a batch of $N$ noisy versions through the model, the time it takes to classify a sample in a robust manner using our method scales linearly with $N$ compared to the same model undefended.
The rest of our method has negligible overhead.
At training time, we essentially have to do perform the same operation over the training dataset, which, depending on its size and the number of desired noise sources, can take a while.
For a given model and dataset, this has to be performed only once however and the computed statistics can then be stored.

\subsection{Logistic classifier for reclassification.}
Instead of selecting class $z$ according to Eq.~(\ref{eq:barF}), we found that training a simple logistic classifier that gets as input all the $K\!-\!1$ Z-scores $\bar g_{y,z}(x)$ for $z \in \{1,..,K\}\backslash y$ can further improve classification accuracy, especially in cases where several Z-scores are comparably far above the threshold.
Specifically, for each class label $y$, we train a separate logsitic regression classifier $C_y$ such that if a sample $x$ of predicted class $y$ is detected as adversarial, we obtain the corrected class label as $z = C_y(x)$. These classifiers are trained on the same training data that is used to collect the statistics for detection.
Two points are worth noting: First, as the classifiers are trained using adversarial samples from a particular adversarial attack model, they might not be valid for adversarial samples from other attack models.
However, we confirm experimentally that our classifiers (trained using PGD) do generalize well to other attacks.
Second, building a classifier in order to protect a classifier might seem tautological, because this metaclassifier could now become the target of an adversarial attack itself.
However, this does not apply in our case, as the inputs to the metaclassifier are (i) low-dimensional (there are just $K\!-\!1$ weight-difference alignments for any given sample), (ii) based on sampled noise and therefore random variables and (iii) the classifier itself is shallow.
All of these make it much harder to specifically attack the corrected classifier.
In Section~\ref{sec:counterattack} we show that our method performs reasonably well even if the adversary is fully aware (has perfect knowledge) of the defense.

\subsection{Additional results mentioned in the main text.}

\begin{figure}[h!]
\centering
	\includegraphics[width=0.9\linewidth]{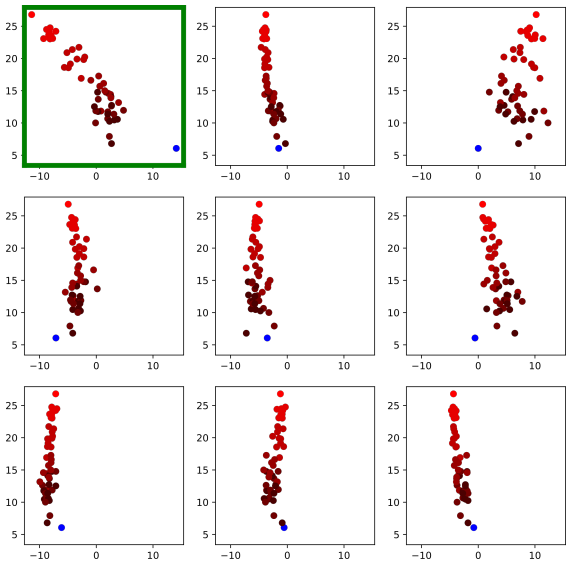}
	\caption{Noise-induced change of logit scores $f_y$ (on the vertical axis) and $f_z$ (on the horizontal axis). Different plots correspond to different classes $z \!\in\! \{1,..,K\}\backslash y$. 
The light red dot shows an adversarially perturbed example $x$ without noise. 
The other red dots show the adversarially perturbed example with added noise.
Color shades reflect noise magnitude: light $=$ small, dark $=$ large magnitude. 
The light blue dot indicates the corresponding natural example without noise. 
The candidate class $z$ in the upper-left corner is selected. See  Figure~\ref{fig:revert} for an explanation.}
\label{fig:weightdifferencealignment}
\end{figure}

\begin{figure}[h]
\begin{tabular}{@{}l@{}c@{\hspace{5pt}}l@{}c@{}} 
	& \adjincludegraphics[width=0.47\linewidth , trim={{0.05\width} {0.08\height} {0.03\width} {0.07\height}},clip]{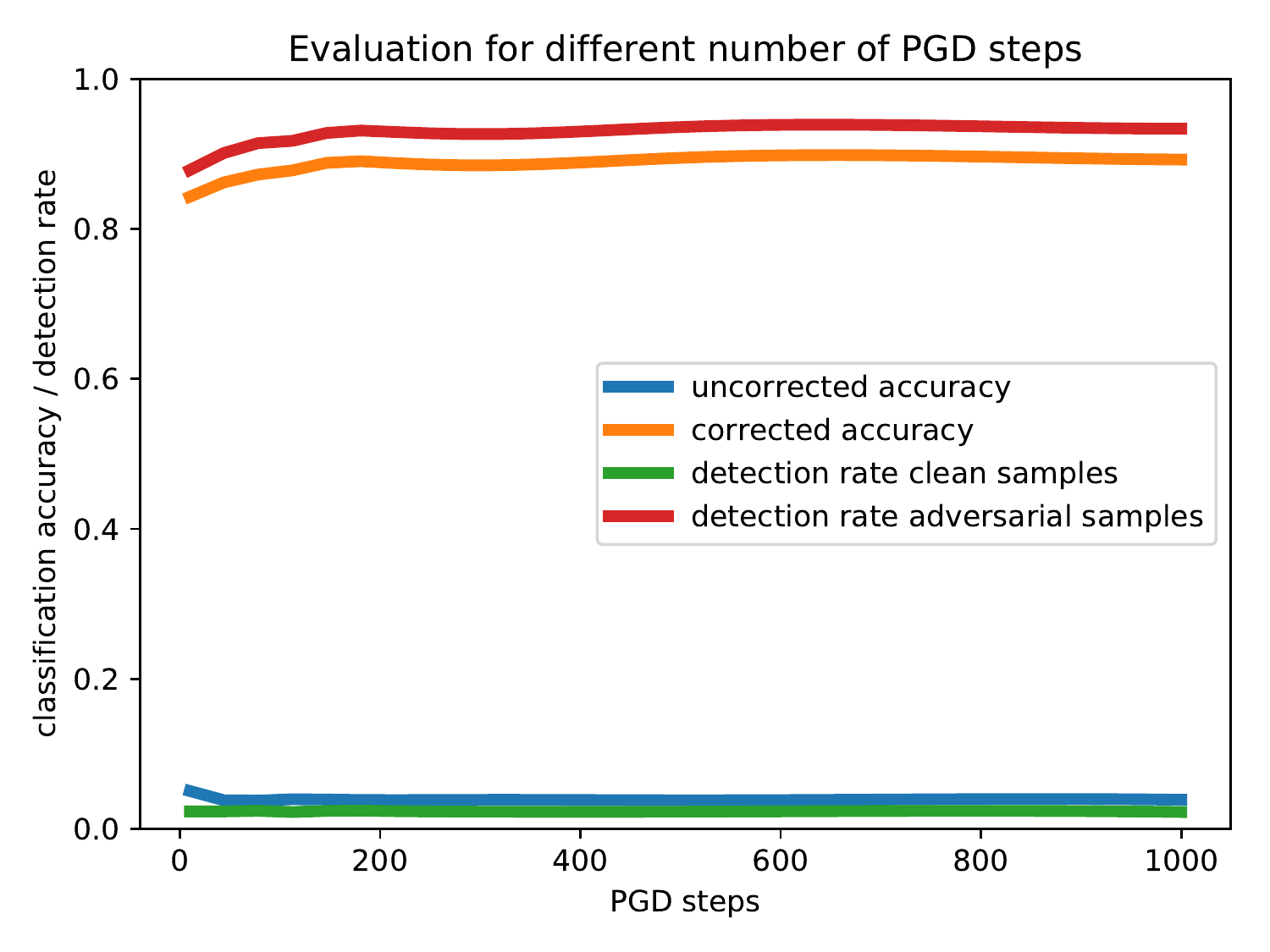} & \hspace{5pt}
	\begin{turn}{90}\hspace{25pt} {$\scriptscriptstyle\langle \triangle\phi, \triangle w \rangle$} \end{turn} & \adjincludegraphics[width=0.47\linewidth , trim={{0.055\width} {0.07\height} {0.028\width} {0.07\height}},clip]{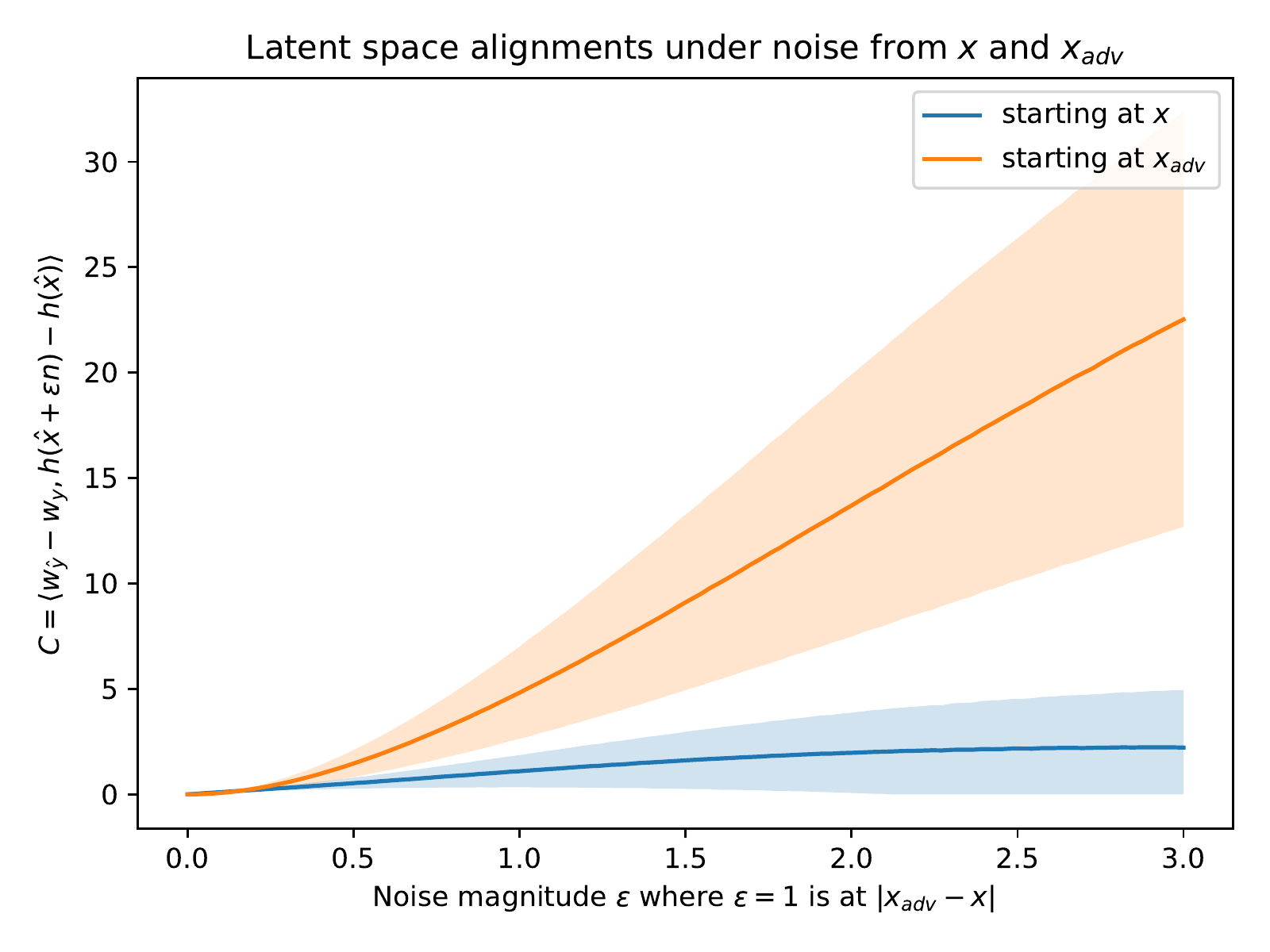}  \\[-3pt]
	&\,\, $\scriptstyle {\rm iterations}$ & & $\scriptstyle t$ \\[-8pt]
\end{tabular}
\caption{(Left) Detection rates and accuracies vs.\ number of PGD iterations.
(Right) Noise-induced weight-difference alignment along $x^*+t\eta$ and $x+t\eta$ respectively. 
For the adversarial example, the alignment with the weight-difference vector between the true and adversarial class is shown. 
For the natural example, the largest alignment with any weight-difference vector is shown. 
}
\vspace{-3mm}
\label{fig:attackiterations}
\label{fig:robustnesstonoise}
\end{figure}

\begin{figure}[h!]
\centering
\begin{tabular}{@{}l@{}c@{\hspace{5pt}}l@{}c@{}}  
	\begin{turn}{90}\hspace{24pt} {$\scriptscriptstyle F_y(x+t\eta)$} \end{turn} & \adjincludegraphics[width=0.437\linewidth , trim={{0.07\width} {0.06\height} {0.1\width} {0.11\height}},clip]{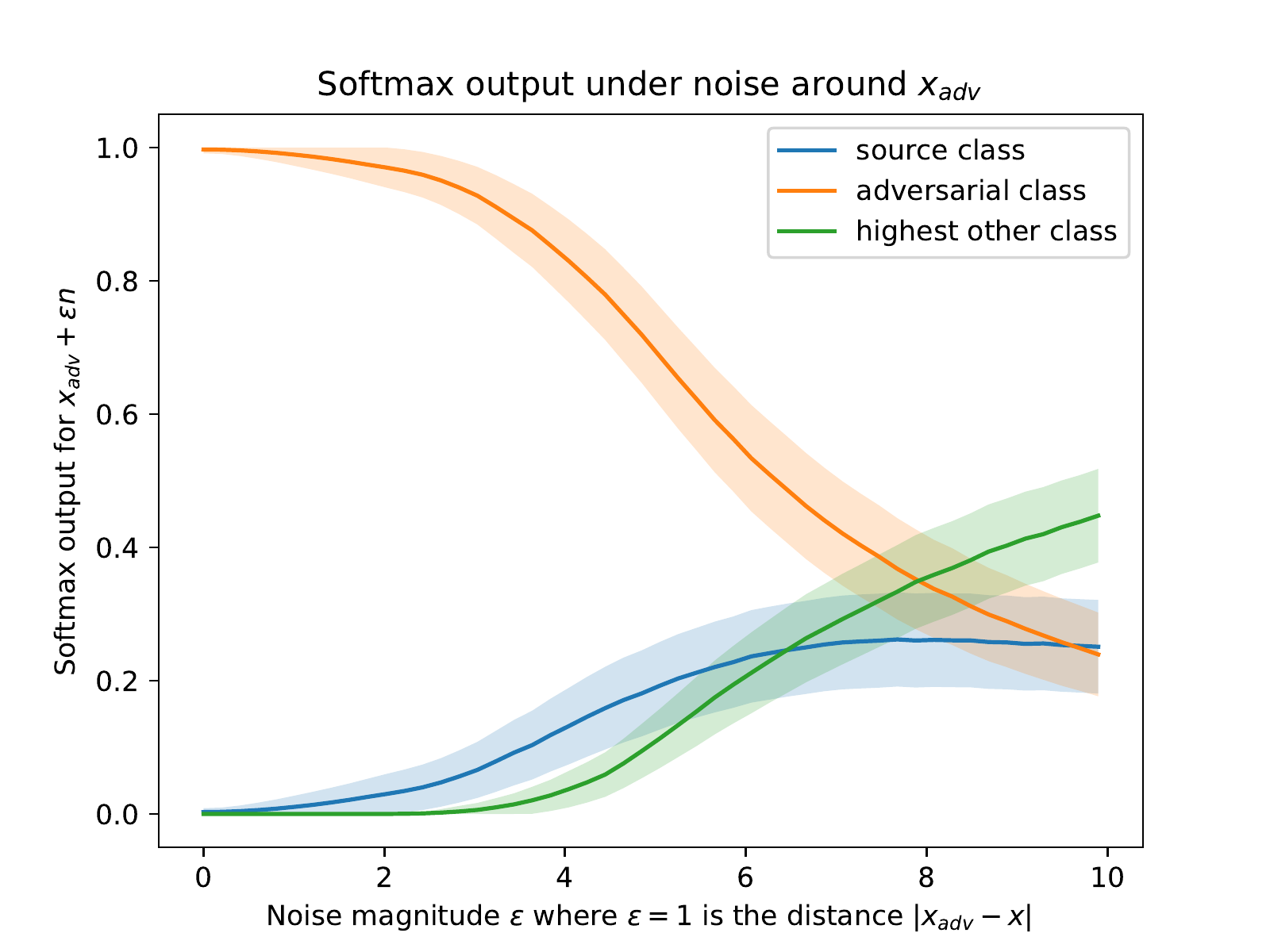} & \hspace{5pt}
	\begin{turn}{90}\hspace{21pt} {$\scriptscriptstyle F_y(x^*+t\Delta x)$} \end{turn} & \adjincludegraphics[width=0.437\linewidth , trim={{0.07\width} {0.06\height} {0.1\width} {0.11\height}},clip]{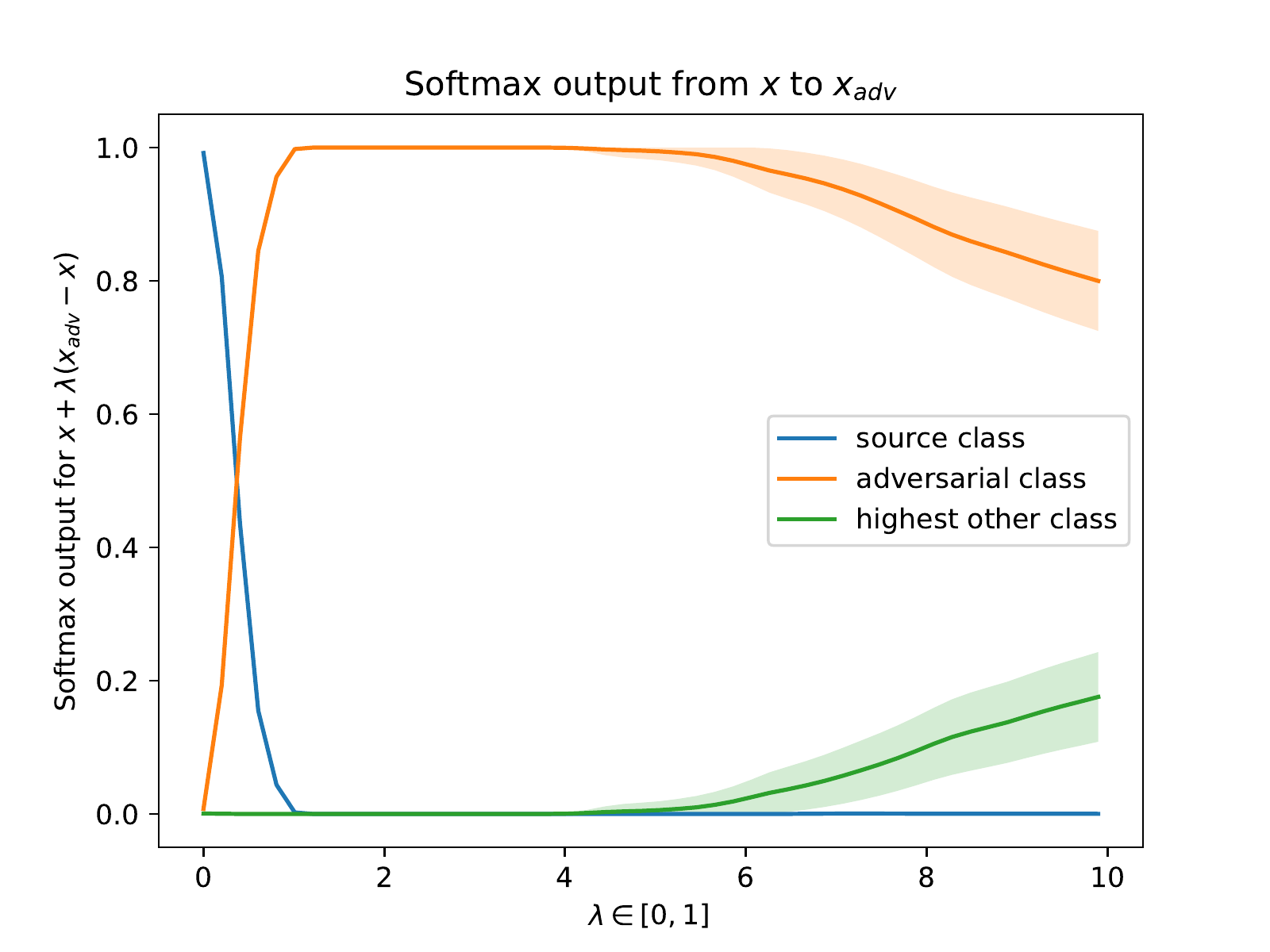} \\[-3pt]
	& $\scriptstyle t$ & & $\scriptstyle t$ \\[-8pt]
\end{tabular}
\caption{
(Left) Softmax predictions $F_y(x+t\eta)$ when adding random noise to the adversarial example. 
(Right) Softmax predictions $F_y(x^*+t\Delta x)$ along the ray from natural to adversarial example and beyond. 
For the untargeted attack shown here, the probability of the source class stays low, even at $t = 10$. 
}
\label{fig:softmaxpredictions}
\end{figure}

\begin{table}[h!]
	\centering
		\caption{Proximity to nearest neighbor. 
		The table shows the ratio of the `distance between the adversarial and the corresponding unperturbed example' to the `distance between the adversarial example and the nearest other neighbor (in either training or test set)', 
		i.e.\ $||x - x^*||_2 / ||x - x^{\rm nn}||_2$.
	}\label{table:ratioofdistances}
	\vskip 0.15in
	\begin{sc}
		\scriptsize
	\begin{tabular}{l c c c}
		 PGD & $\epsilon_\infty\!=\!2$ & $\epsilon_\infty\!=\!4$ & $\epsilon_\infty\!=\!8$ \\[2pt]
		\hline\\[-8pt]
		$L^\infty$ & $0.021 \pm 0.005$ & $0.039 \pm 0.010$ & $0.075 \pm 0.018$ \\[2pt]
		$L^2$      &  $0.023 \pm 0.006$ & $0.043 \pm 0.012$ & $0.088 \pm 0.019$
	\end{tabular}
	\end{sc}
 
\end{table}

\subsection{ROC Curves.}
Figure~\ref{fig:roc} shows how our method performs against a PGD attack under different settings of thresholds $\tau$.
\begin{figure}[h]
	\centering
	\begin{tabular}{@{}l@{}c@{\hspace{5pt}}l@{}c@{}} 
		& {\sc WResNet} & & {\sc WResNet} \\
		\begin{turn}{90}\hspace{5pt} {\tiny accuracy on pgd samples} \end{turn} & \adjincludegraphics[width=0.45\linewidth , trim={{0.06\width} {0.085\height} {0.02\width} {0.075\height}},clip]{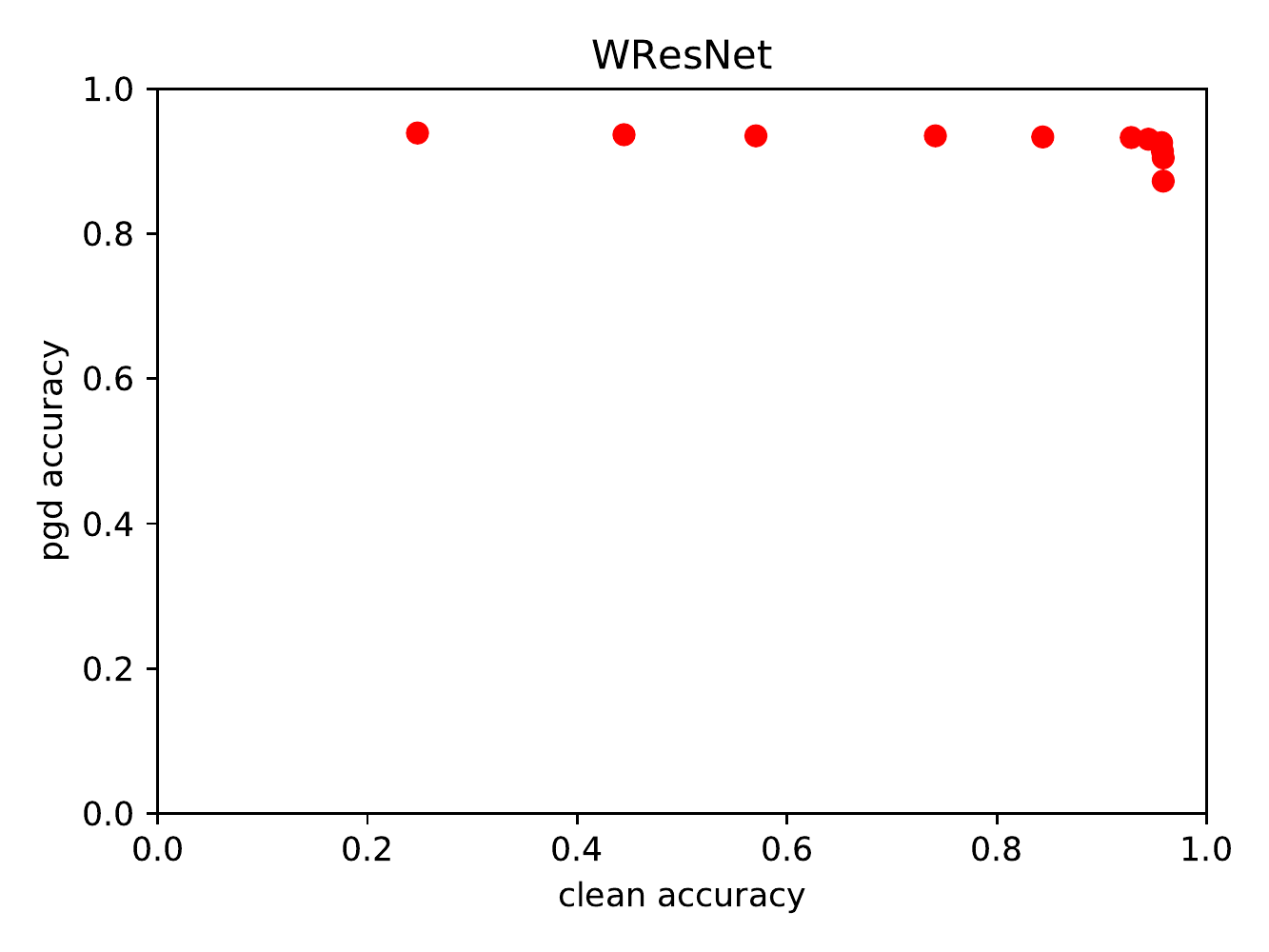} &
		\begin{turn}{90}\hspace{4pt} {\tiny true positive detection rate} \end{turn} & \adjincludegraphics[width=0.45\linewidth , trim={{0.06\width} {0.085\height} {0.02\width} {0.075\height}},clip]{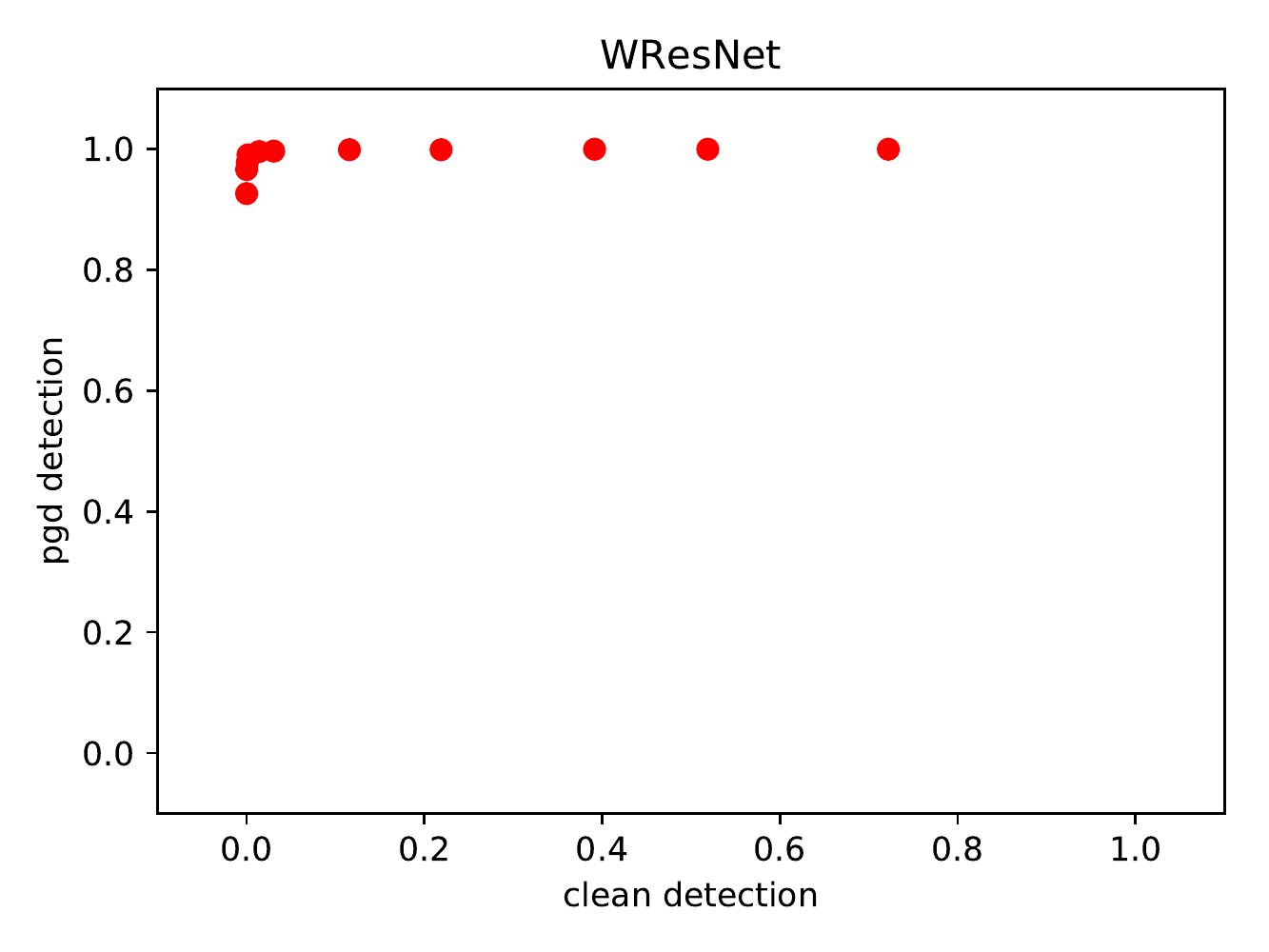}\\[-5pt]
		& \,\,{\tiny accuracy on clean samples} & & {\tiny false positive detection rate} \\[-5pt]
		\\
		& {\sc CNN7} & & {\sc CNN7} \\
		\begin{turn}{90}\hspace{5pt} {\tiny accuracy on pgd samples} \end{turn} & \adjincludegraphics[width=0.45\linewidth , trim={{0.06\width} {0.085\height} {0.02\width} {0.075\height}},clip]{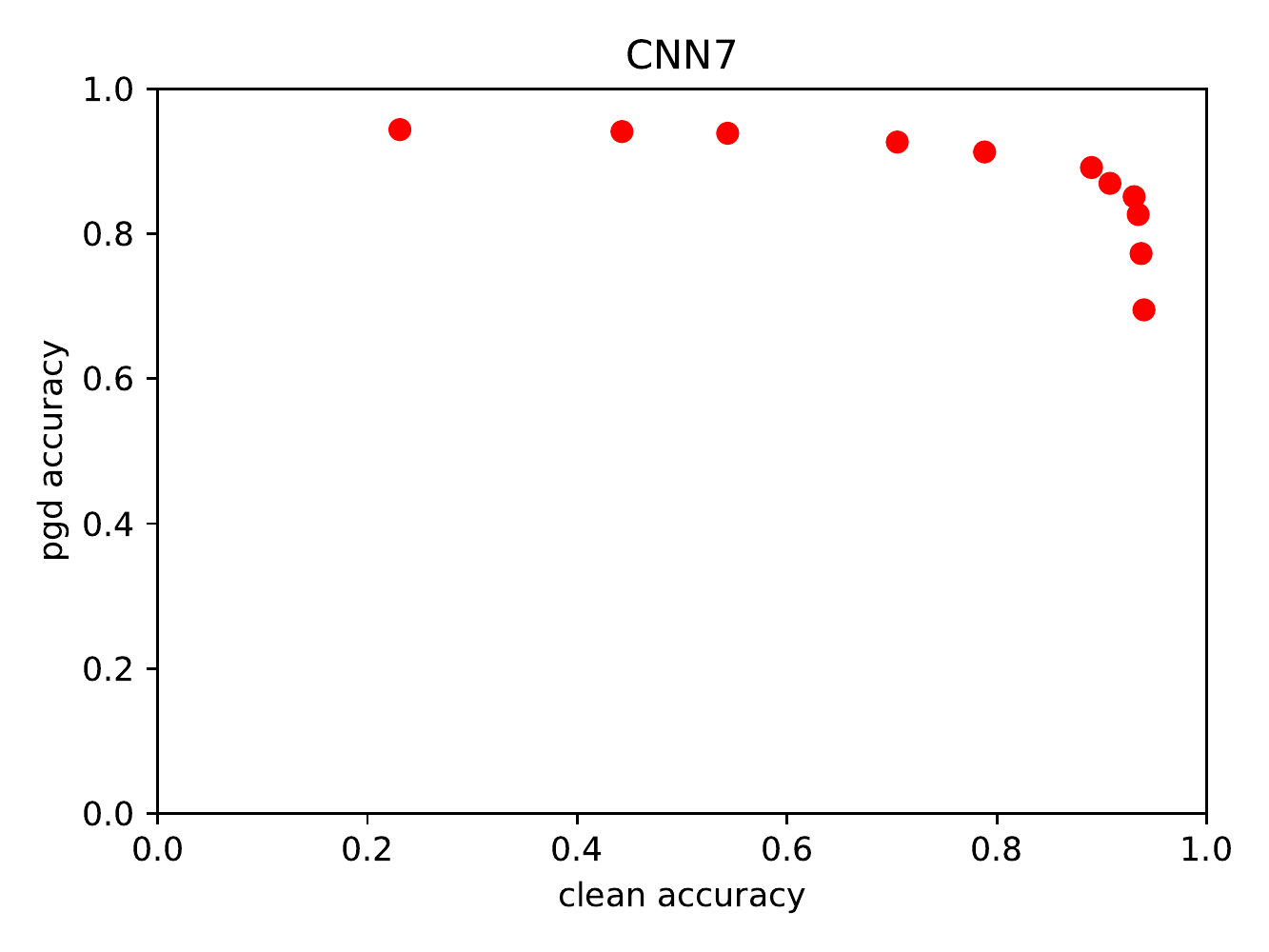} &
		\begin{turn}{90}\hspace{4pt} {\tiny true positive detection rate} \end{turn} & \adjincludegraphics[width=0.45\linewidth , trim={{0.06\width} {0.085\height} {0.02\width} {0.075\height}},clip]{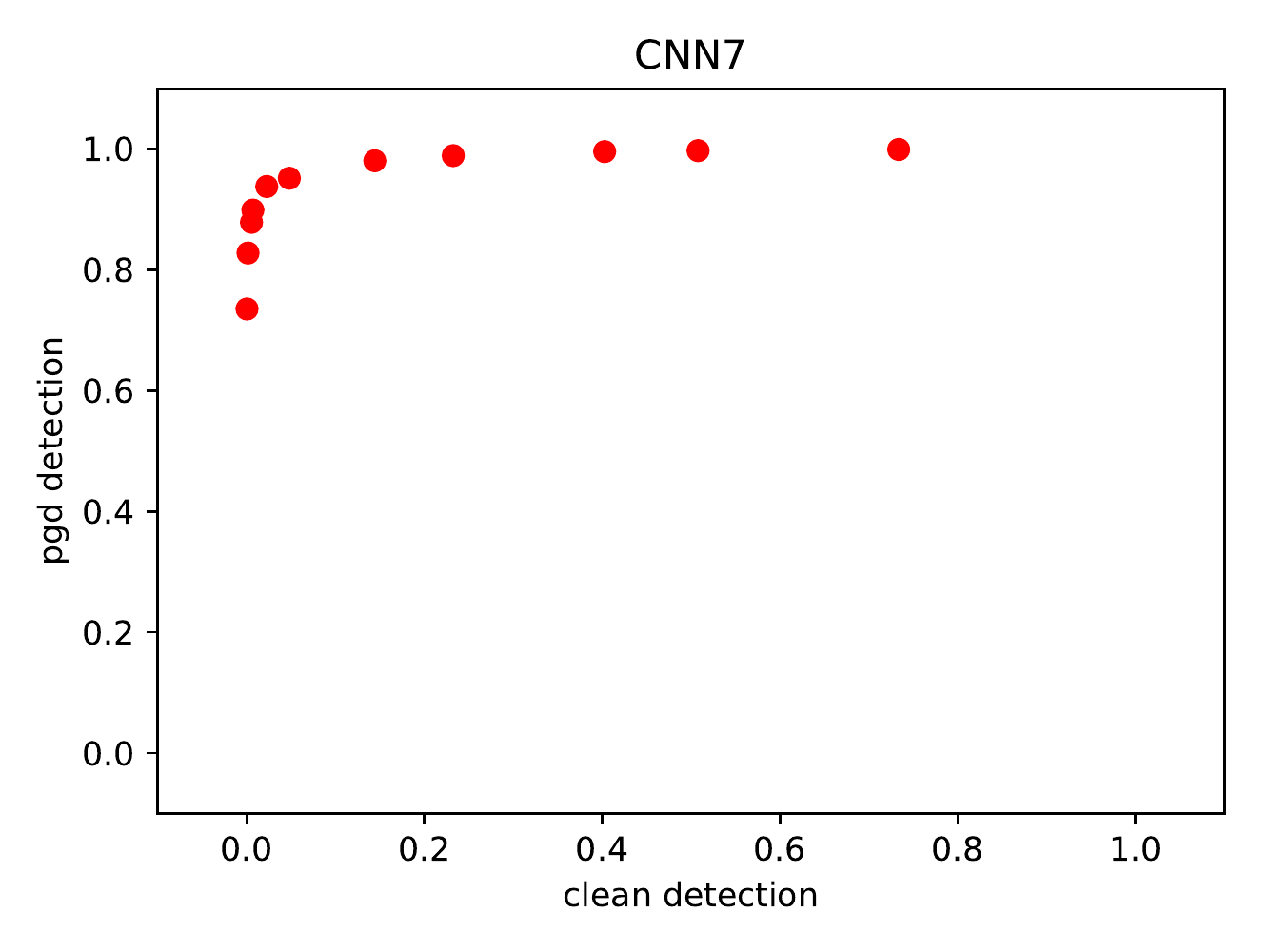}\\[-5pt]
		& \,\,{\tiny accuracy on clean samples} & & {\tiny false positive detection rate} \\[-5pt]
		\\
		& {\sc CNN4} & & {\sc CNN4} \\
		\begin{turn}{90}\hspace{5pt} {\tiny accuracy on pgd samples} \end{turn} & \adjincludegraphics[width=0.45\linewidth , trim={{0.06\width} {0.085\height} {0.02\width} {0.075\height}},clip]{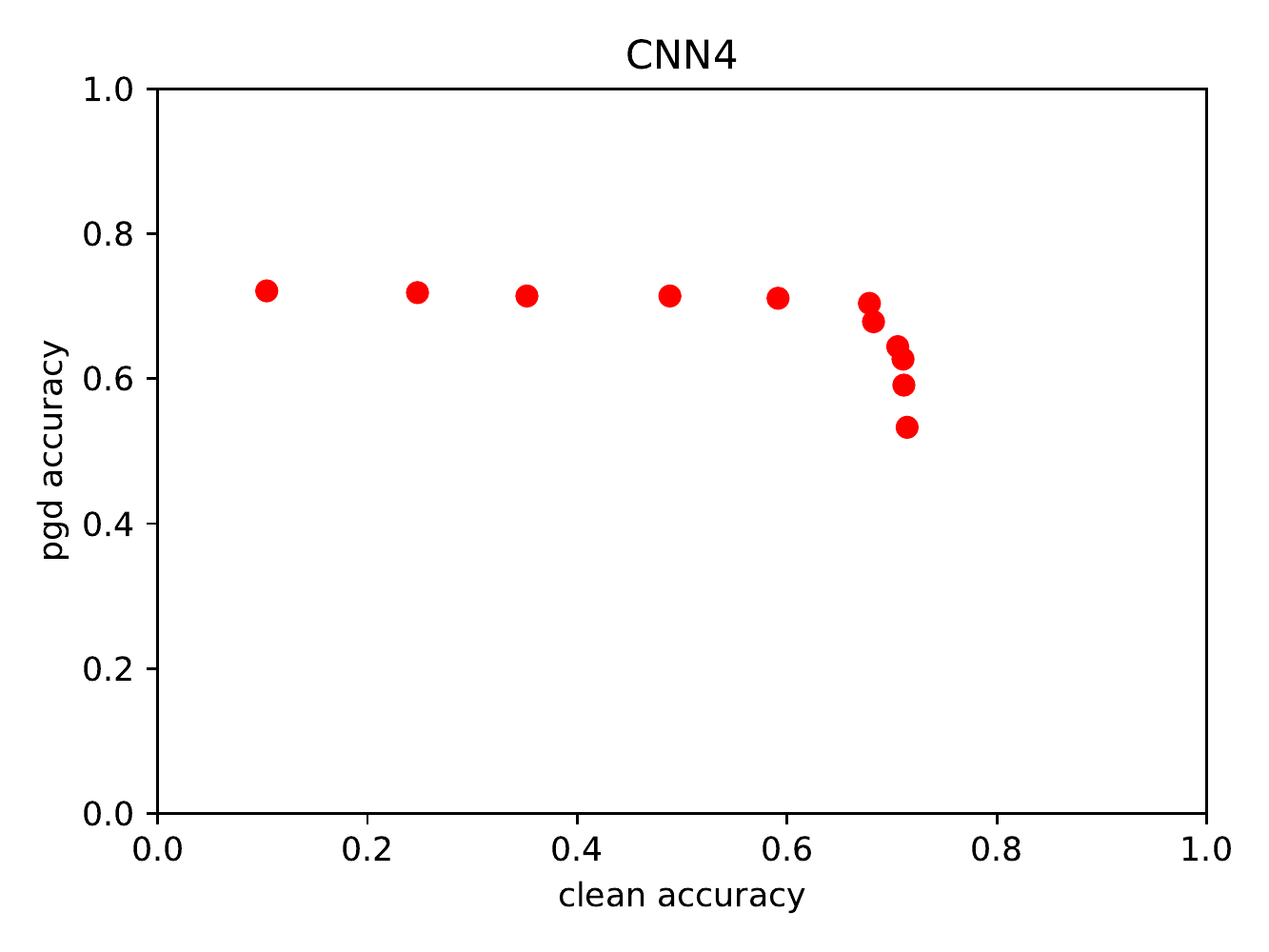} &
		\begin{turn}{90}\hspace{4pt} {\tiny true positive detection rate} \end{turn} & \adjincludegraphics[width=0.45\linewidth , trim={{0.06\width} {0.085\height} {0.02\width} {0.075\height}},clip]{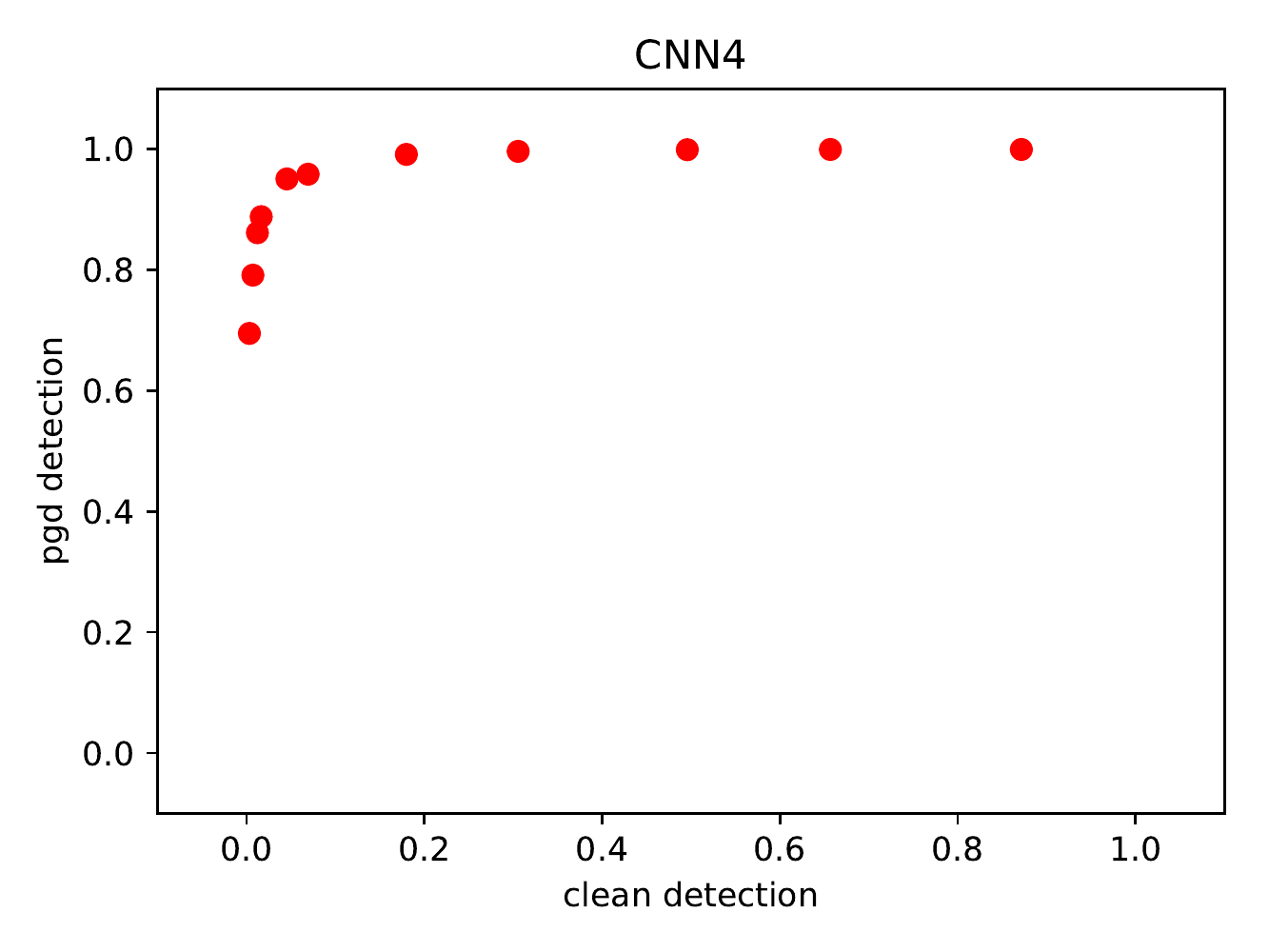}\\[-5pt]
		& \,\,{\tiny accuracy on clean samples} & & {\tiny false positive detection rate} \\[-5pt]
	\end{tabular}
	\caption{ROC-curves. Test set accuracies and detection rates on clean and PGD-perturbed samples for a range of thresholds $\tau$ on CIFAR10.}
	\label{fig:roc}
\end{figure}

\end{document}